\documentclass[10pt,twocolumn,letterpaper]{article}

\usepackage{cvpr}              %
\usepackage[utf8]{inputenc}
\usepackage[T1]{fontenc}          
\usepackage{url}           
\usepackage{booktabs}       
\usepackage{amsfonts}       
\usepackage{nicefrac}       
\usepackage{microtype}      
\usepackage{xcolor}

\usepackage{amsmath}
\usepackage{booktabs} 
\usepackage{multirow} 
\usepackage{pifont} 
\usepackage{graphicx} 
\usepackage{wrapfig} 
\usepackage{adjustbox}
\usepackage{multirow}
\usepackage{dsfont}
\usepackage[table]{xcolor}
\usepackage[accsupp]{axessibility}

\newcommand{\cL}{\mathcal{L}}

\newcommand{\figref}[1]{Fig.~\ref{#1}}
\newcommand{\secref}[1]{Sec.~\ref{#1}}

\newcommand{\eqnref}[1]{Eq.~\eqref{#1}}
\newcommand{\tabref}[1]{Tab.~\ref{#1}}

\makeatletter
\DeclareRobustCommand\onedot{\futurelet\@let@token\@onedot}
\def\@onedot{\ifx\@let@token.\else.\null\fi\xspace}
\def\eg{e.g\onedot} 
\def\ie{i.e\onedot}

\makeatother

\newcommand{\boldparagraph}[1]{\vspace{0.0cm}\noindent{\bf #1.}}

\definecolor{darkgreen}{rgb}{0,0.7,0}
\definecolor{darkyellow}{rgb}{0.8,0.8,0}
\definecolor{bittersweet}{rgb}{1.0, 0.44, 0.37}
\definecolor{amber}{rgb}{1.0, 0.49, 0.0}
\definecolor{lgray}{rgb}{0.83,0.83,0.83}

\definecolor{color_unlabled}{rgb}{0.0,0.0,0.0}
\definecolor{color_vehicle}{rgb}{0.0,0.0,0.56}
\definecolor{color_road}{rgb}{0.5,0.25,0.5}
\definecolor{color_redlight}{rgb}{1.0,0.0,0.0}
\definecolor{color_person}{rgb}{0.859,0.078,0.234}
\definecolor{color_roadline}{rgb}{0.613,0.914,0.195}
\definecolor{color_sidewalk}{rgb}{0.953,0.137,0.906}

\definecolor{ellisred}{rgb}{0.87,0.44,0.38} %
\definecolor{ellisgreen}{rgb}{0.69,0.90,0.52} %
\definecolor{elliscyan}{rgb}{0.29,0.77,0.74} %
\definecolor{ellisorange}{rgb}{0.89,0.55,0.28} %
\definecolor{ellisblue}{rgb}{0.41,0.61,0.86} %

\definecolor{Tab0}{HTML}{1F77B4}
\definecolor{Tab1}{HTML}{ff7f0e}
\definecolor{Tab2}{HTML}{2ca02c}
\definecolor{Tab3}{HTML}{d62728}
\definecolor{Tab4}{HTML}{9467bd}
\definecolor{Tab5}{HTML}{8c564b}
\definecolor{Tab6}{HTML}{e377c2}
\definecolor{Tab7}{HTML}{7f7f7f}
\definecolor{Tab8}{HTML}{bcbd22}
\definecolor{Tab9}{HTML}{17becf}

\definecolor{Tabx0}{HTML}{4e79a7}
\definecolor{Tabx1}{HTML}{f28e2b}
\definecolor{Tabx2}{HTML}{e15759}
\definecolor{Tabx3}{HTML}{76b7b2}
\definecolor{Tabx4}{HTML}{59a14f}
\definecolor{Tabx5}{HTML}{edc948}
\definecolor{Tabx6}{HTML}{b07aa1}
\definecolor{Tabx7}{HTML}{ff9da7}
\definecolor{Tabx8}{HTML}{9c755f}
\definecolor{Tabx9}{HTML}{bab0ac}

\definecolor{Cx0}{HTML}{4e79a7}
\definecolor{Cx1}{HTML}{f28e2b}
\definecolor{Cx2}{HTML}{e15759}
\definecolor{Cx3}{HTML}{76b7b2}
\definecolor{Cx4}{HTML}{59a14f}
\definecolor{Cx5}{HTML}{edc948}
\definecolor{Cx6}{HTML}{b07aa1}
\definecolor{Cx7}{HTML}{ff9da7}
\definecolor{Cx8}{HTML}{9c755f}
\definecolor{Cx9}{HTML}{bab0ac}

\definecolor{kitti_ground_color}{HTML}{510051}
\definecolor{kitti_terrain_color}{HTML}{98fb98}
\definecolor{kitti_sidewalk_color}{HTML}{f423e8}
\definecolor{kitti_parking_color}{HTML}{faaaa0}
\definecolor{kitti_road_color}{HTML}{804080}
\definecolor{kitti_vegetation_cuboid_color}{HTML}{6b8e23}
\definecolor{kitti_vegetation_ellipsoid_color}{HTML}{6b8e23}
\definecolor{kitti_vehicle_big_color}{HTML}{003c64}
\definecolor{kitti_vehicle_small_color}{HTML}{00008e}
\definecolor{kitti_two_wheelers_color}{HTML}{770b20}
\definecolor{kitti_human_color}{HTML}{dc143c}
\definecolor{kitti_construction_big_color}{HTML}{464646}
\definecolor{kitti_construction_small_color}{HTML}{66669c}
\definecolor{kitti_pole_color}{HTML}{999999}
\definecolor{kitti_traffic_control_color}{HTML}{faaa1e}
\definecolor{kitti_object_color}{HTML}{0080c0}

\usepackage{pifont}
\newcommand{\cmark}{\ding{51}}%
\newcommand{\xmark}{\ding{55}}%
\newcommand{\best}[1]{\cellcolor{red!30}#1}
\newcommand{\second}[1]{\cellcolor{orange!30}#1}
\newcommand{\third}[1]{\cellcolor{yellow!30}#1}

\DeclareRobustCommand{\bestcap}[1]{\colorbox{red!30}{#1}}
\DeclareRobustCommand{\secondcap}[1]{\colorbox{orange!30}{#1}}
\DeclareRobustCommand{\thirdcap}[1]{\colorbox{yellow!30}{#1}}

\usepackage{tabularx}

\definecolor{cvprblue}{rgb}{0.21,0.49,0.74}
\usepackage[pagebackref,breaklinks,colorlinks,allcolors=cvprblue]{hyperref}

\def\paperID{} %
\def\confName{CVPR}
\def\confYear{2026}

\title{PrITTI: Primitive-based Generation of \\ 
Controllable and Editable 3D Semantic Urban Scenes}

\author{
Christina Ourania Tze$^{1}$ \quad
Daniel Dauner$^{1,4}$ \quad
Yiyi Liao$^{2}$ \quad
Dzmitry Tsishkou$^{3}$ \quad 
Andreas Geiger$^{1,4}$
\vspace{0.1cm}
\\
$\small ^{1}$University of T{\"u}bingen, T{\"u}bingen AI Center \quad
$\small ^{2}$Zhejiang University \quad
$\small ^{3}$Noah’s Ark Lab, Huawei \\
$\small ^{4}$KE:SAI - Kyutai ELLIS Scalable Autonomous Intelligence \\
}

\begin{document}

\twocolumn[{%
\renewcommand\twocolumn[1][]{#1}%
\begin{center}
\maketitle
\vspace{-5mm}
\centerline{\includegraphics[width=\textwidth]{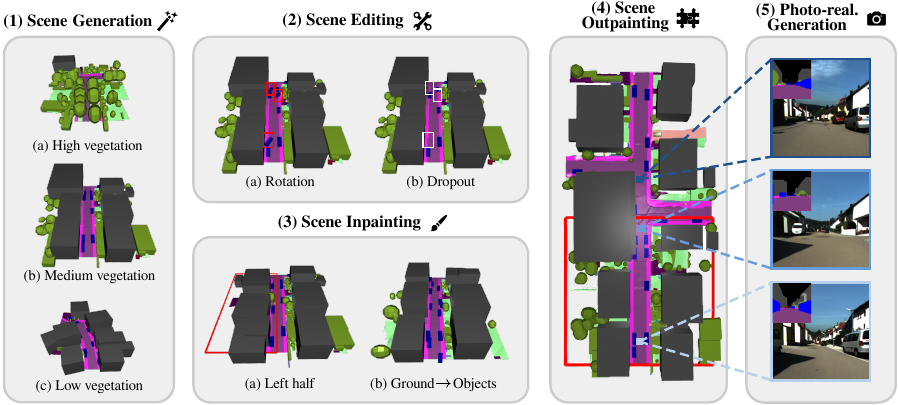}}
   \vspace{-0.1cm}
   \captionof{figure}{
   \textbf{PrITTI} generates (1) high-quality, controllable 3D semantic urban scenes %
   in a compact primitive-based representation using a latent diffusion model. Starting from a generated scene (\eg middle sample), we demonstrate downstream applications including (2) scene editing, (3) inpainting,
    (4) outpainting, and (5) photo-realistic street view synthesis.
   }
    \label{fig:teaser}
    \vspace{-0.1cm}
\end{center}%
}]
\begin{abstract}
Existing approaches to 3D semantic urban scene generation predominantly rely on voxel-based representations, which are bound by fixed resolution, challenging to edit, and memory-intensive in their dense form. In contrast, we advocate for a primitive-based paradigm where urban scenes are represented using compact, semantically meaningful 3D elements that are easy to manipulate and compose. To this end, we introduce PrITTI, a latent diffusion model that leverages vectorized object primitives and rasterized ground surfaces for generating diverse, controllable, and editable 3D semantic urban scenes. This hybrid representation yields a structured latent space that facilitates object- and ground-level manipulation. Experiments on KITTI-360 show that primitive-based representations unlock the full capabilities of diffusion transformers, achieving state-of-the-art 3D scene generation quality with lower memory requirements, faster inference, and greater editability than voxel-based methods. Beyond generation, PrITTI supports a range of downstream applications, including scene editing, inpainting, outpainting, and photo-realistic street-view synthesis. The source code and more results can be found at \url{https://raniatze.github.io/pritti/}.
\vspace{-0.3cm}
\end{abstract}
    
\section{Introduction}
\label{sec:introduction}
Generating large-scale, semantically structured 3D urban scenes is essential for digital world modeling and autonomous driving simulation. Such virtual worlds enable safety validation, rare-scenario testing, and sensor data synthesis for perception and planning. However, constructing them typically requires expert knowledge and extensive manual effort, making the process costly and time-consuming. Although some structured 3D datasets exist, their scarcity, particularly for outdoor settings, restricts diversity and yields repetitive scene patterns. These limitations motivate generative approaches capable of automatically producing diverse and realistic urban environments. To be practically useful, such methods should additionally support controllability and flexible editing. The extent to which these capabilities can be achieved largely depends on the chosen underlying representation. An ideal representation should be memory-efficient, compositional, and support localized, intuitive edits. Most existing approaches model 3D urban environments using voxel grids~\cite{Lee2024CVPR, Zheng2024ARXIV} or hierarchical voxel structures~\cite{Ren2024CVPR,Liu2024ECCV}, which suffer from three fundamental limitations. %
First, due to cubic scaling with resolution, they are 
\textit{memory-inefficient} and prohibitively expensive for large environments. Second, their fixed spatial resolution leads to \textit{limited scene detail}. For instance, tall elements of a scene (\eg buildings or trees) often appear truncated or distorted. Third, voxel grids are \textit{difficult to manipulate}. Even simple edits, like translating a vehicle, require identifying and updating all related voxels and then filling the vacated space, a process which is complicated by ambiguous object-background boundaries (\eg car vs. road). These challenges highlight the need for a more compact, flexible, and object-centric scene representation.

In this work, we advocate for primitives as a compelling alternative for scalable 3D semantic scene synthesis, owing to their compositional and parsimonious nature. Despite their extensive use in shape abstraction~\cite{Tulsiani2017CVPRb, Zou2017ICCVb, Li2017TG, Mo2019TG} and scene parsing~\cite{Gupta2010ECCV}, their potential in generative settings, especially for synthesizing 3D semantic urban scenes, remains largely underexplored. Motivated by this gap, we present PrITTI, the first framework to use coarse 3D bounding primitives for generating controllable and editable 3D semantic scenes in urban environments. While recent approaches~\cite{Yang2023ICCV, Lu2024ARXIV, Bahmani2023ICCV, Po2024THREEDV, Xu2023CVPR, Yang2024NeurIPS, Chen2025ARXIV} employ abstract semantic layouts, these are typically hand-crafted or pre-defined inputs rather than generated. %
Our method adopts the common primitive-based scene representations~\cite{Caesar2021CVPR,Caesar2021CVPRW,Argoverse22021NeurIPS,Liao2022PAMI}, modeling objects as 3D cuboids (\eg buildings, vehicles) or ellipsoids (\eg trees) and ground surfaces as extruded 2D polygons of varying shapes (\eg road, sidewalk).
Although cuboids and ellipsoids allow consistent parameterization, the arbitrary geometry of polygons makes uniform encoding impractical. We address this via a hybrid representation: ground elements are rasterized, while objects are modeled as parameterized primitives (cuboids and ellipsoids). Training proceeds in two stages: a variational autoencoder (VAE)~\cite{Kingma2014ICLR} first encodes 3D layouts into compact 2D latents, followed by a latent diffusion model (LDM)~\cite{Rombach2022CVPR} trained on discrete primitive density labels for controllable scene generation. Our LDM supports downstream tasks such as scene inpainting and outpainting without fine-tuning (see \figref{fig:teaser}). This is achieved via a latent manipulation mechanism~\cite{Lugmayr2022CVPR} that allows localized edits while preserving global structure. A key component of our approach is its disentangled latent structure, where ground and object elements are encoded in two separate channel groups. This design enables further applications (\eg ground-conditioned object generation) without extra supervision. Additionally, our instance-level primitive representation supports intuitive editing operations (\eg translation, scaling, rotation). %
Finally, we introduce a Cholesky-based~\cite{Cholesky1924} parameterization of 3D object orientation and size, offering greater numerical stability over quaternion encodings, as evidenced by our experiments. 

To summarize, our method introduces: \textbf{(i)} the first primitive-based framework for large-scale urban 3D scene generation at an abstract, semantic level, \textbf{(ii)} a disentangled latent representation enabling controllable manipulation, and \textbf{(iii)} a stable encoding of 3D object orientation and size based on Cholesky decomposition. Experiments demonstrate that PrITTI achieves competitive reconstruction with lower memory requirements and delivers superior generation quality, diversity, and editability compared to voxel-based baselines.

\section{Related Work}
\label{sec:related_work}
\boldparagraph{Primitive-based Representation}
Primitives have been explored since the early days of computer vision~\cite{Roberts1963PHD, Binford1975CSC, Brooks1979IJCAI, Biederman1985, Pentland1986AAAI} as compact, compositional building blocks for objects and scenes. With deep learning, these ideas re-emerged for decomposing 3D shapes into semantically meaningful parts or recovering object structure from RGB or RGB-D inputs. Most works represent objects using simple cuboids~\cite{Tulsiani2017CVPRb, Zou2017ICCVb, Niu2018CVPR, Li2017TG, Mo2019TG}, while others adopt more expressive primitive types such as superquadrics~\cite{Paschalidou2019CVPR, Paschalidou2020CVPR}, convexes~\cite{Deng2020CVPR}, axis-aligned Gaussians~\cite{Genova2019ICCV}, and spheres~\cite{Hao2020CVPR}. Recent part-aware generative approaches like SPAGHETTI~\cite{Hertz2022SIGGRAPH} and SALAD~\cite{Koo2023ICCV} decompose objects into primitive-based sub-parts (\eg legs, seat of a chair) for part-level manipulation. Beyond individual objects, few recent methods decompose indoor~\cite{Vavilala2024ARXIV, Vavilala2023ICCV, Vavilala2023ARXIV} or outdoor scenes~\cite{Gupta2010ECCV} into simple geometric shapes. While~\cite{Li2017SIGGRAPH, Mo2019SIGGRAPH, Zou2017ICCVb, Hertz2022SIGGRAPH, Koo2023ICCV} demonstrate primitive generation at the object level, the majority of prior approaches instead use primitives for shape abstraction or scene parsing. In contrast, we leverage parameterized primitives, each corresponding to a distinct object instance, for abstract 3D semantic layout generation, enabling direct scene-level operations (\eg object arrangement), which remain beyond the scope of prior object-centric methods.

\boldparagraph{3D Scene Generation} %
Recent works tackle 3D scene generation from different perspectives, aiming to synthesize either detailed geometry or realistic appearances. 
Geometry-focused methods~\cite{Bokhovkin2024ARXIV,Wu2024TOG,Meng2025CVPR} produce high-fidelity meshes but lack semantic and compositional structure. Early appearance-oriented approaches~\cite{Fridman2023NeurIPS,Hoellein2023ICCV,Song2023ARXIV} update explicit meshes to generate immersive 3D scenes from text. Other methods adopt volumetric scene representations, optimizing NeRFs~\cite{Zhang2023ARXIV} or 3D Gaussians~\cite{Yang2025CVPR,Shriram20243DV,Ouyang2023ARXIV} for global consistency and photorealism. More recent frameworks~\cite{Yu2024CVPRb,Mark2025ICCV,Ren2025CVPR,Yu2024ARXIV} introduce 3D-aware caches that guide view generation toward spatio-temporal coherence.  %
Finally, two-stage panoramic methods~\cite{Schwarz2025ICCV,Yang2024SIGGRAPH,Huang2025CVPR,Worldgen2025Github} 
generate 3D scenes from text or single images by first synthesizing a 2D panorama and then lifting it to 3D to optimize Gaussian splats. Despite impressive realism, these appearance-driven approaches represent scenes holistically, without explicit, object-level structure. In contrast, we emphasize a controllable semantic layout over appearance or geometric fidelity, enabling direct instance-level control and high-level manipulation.

\boldparagraph{Semantic Layout Generation} Prior work in this area can be divided into indoor and outdoor approaches. Early indoor methods generate objects sequentially using autoregressive models~\cite{Wang2021THREEDV, Ritchie2019CVPR, Paschalidou2021NEURIPS}, whereas recent approaches~\cite{Wei2023CVPR, Tang2024CVPR} predict all elements jointly using transformers or diffusion models. Conditioning signals include floor plans~\cite{Paschalidou2021NEURIPS}, partial layouts~\cite{Ritchie2019CVPR, Paschalidou2021NEURIPS, Wang2021THREEDV}, text~\cite{Wang2021THREEDV}, and scene graphs~\cite{Luo2020CVPR, Zhai2023NeurIPS, Dhamo2021ICCV, Zhai2024ECCV, Lin2024ICLR}. While effective at room scale, these techniques have not been demonstrated in outdoor settings, which involve %
higher object counts and greater geometric diversity. Early methods in this direction~\cite{Xie2024CVPR, Lin2023ICCV, Deng2023ARXIV, Xie2024ARXIV, Xie2025ARXIV} produce 2.5D layouts by generating pairs of top-down semantic and height maps, then lifting objects into 3D via vertical extrusions of 2D footprints. However, they are limited by the generated maps' resolution and cannot capture complex scene geometry. Recently, voxel grids have been applied for 3D semantic scene generation.~\cite{Lee2024CVPR} and~\cite{Zheng2024ARXIV} compress voxels into latent triplanes to improve efficiency and train unconditional or mask-conditional latent diffusion models, respectively. Hierarchical voxel models~\cite{Ren2024CVPR, Liu2024ECCV} further aim to improve scalability by iteratively refining scenes across multiple resolutions. However, this design requires training multiple VAEs and diffusion models per hierarchy level. Moreover, although sparse voxel representations can significantly reduce memory requirements, \eg as implemented in XCube~\cite{Ren2024CVPR} via custom CUDA operations, voxel-based approaches still inherit the fundamental limitations of voxel grids, including limited editability and fixed spatial resolution. Semantic scene generation has also been applied in generative driving simulators to synthesize traffic layouts, using agent bounding boxes~\cite{Lu2024ICRA, Tan2021CVPR, Jiang2024NeurIPS, Zhou2025ARXIV}, lane graphs~\cite{Mi2021CVPR}, or both~\cite{Sun2024RAL, Rowe2025ARXIV, Chitta2024ECCV}, with the closest to ours being SLEDGE~\cite{Chitta2024ECCV}. Unlike these 2D abstraction-based methods, our approach models 3D scenes with richer primitive types, semantic classes, and flexible ground geometry beyond 2D polylines.

\section{Method}
\label{sec:method}
We introduce PrITTI, a latent diffusion model for controllable and editable generation of 3D primitive-based semantic scene layouts. \figref{fig:01_method} shows an overview of our approach. 

\subsection{3D Semantic Scene Layout Representation}
\label{subsec:3d_scene_layout}
\begin{figure*}[t!]
\centering
 \includegraphics[width=\textwidth]{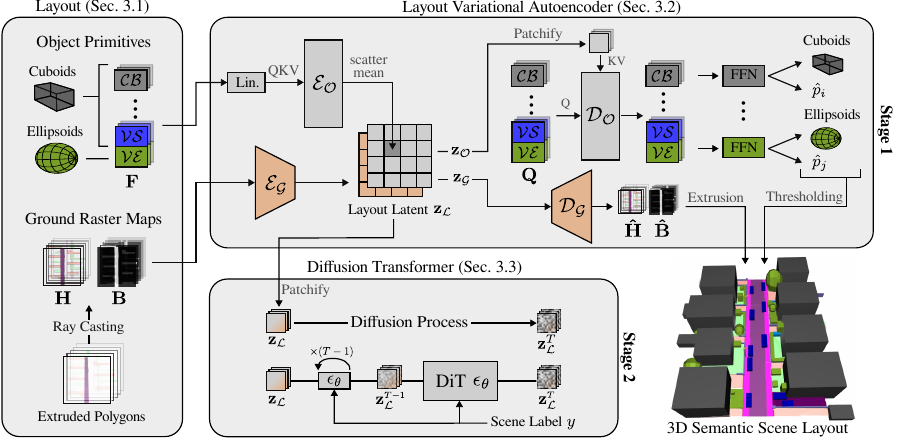}
\vspace{-0.3cm}
\caption{\textbf{Training Overview.} An input 3D semantic layout \( \mathcal{S} \) comprises object primitives, encoded as feature vectors \( \mathbf{F} \), and extruded ground polygons, rasterized into height maps \( \mathbf{H} \) and binary occupancy masks \( \mathbf{B} \) (\secref{subsec:3d_scene_layout}). A layout VAE with separate encoder-decoder pairs for objects (\( \mathcal{E}_\mathcal{O}\)/\(\mathcal{D}_\mathcal{O}\)) and ground (\(\mathcal{E}_\mathcal{G}\)/\(\mathcal{D}_\mathcal{G}\)) first compresses \( \mathcal{S} \) into a structured latent representation \( \mathbf{z}_\mathcal{L} \) (\secref{subsec:lvae}). In the second stage, a diffusion model is trained over this latent space for controllable scene generation (\secref{subsec:dit}). At inference, the diffusion model generates latent codes either unconditionally or conditioned on the scene label \( y\), which are then decoded by the VAE into novel 3D layouts.}
\label{fig:01_method}
\vspace{-0.6cm}
\end{figure*}

An input 3D semantic layout \( \mathcal{S} \) contains 3D cuboids and ellipsoids for object instances and upwards-extruded polygons for ground surfaces. Given their structural differences, we employ distinct representations for each element type:

\boldparagraph{Ground Raster Maps} Ground annotations include five semantic classes (road, sidewalk, parking, terrain, ground) and are defined as 2D polygons extruded into 3D (see \figref{fig:01_method}). Their varying shapes and vertex counts make consistent parameterization difficult. We therefore perform ray casting along the gravity direction to generate bird's eye view (BEV) ground raster maps that are well-suited for convolutional networks. These consist of height maps \( \mathbf{H} \in \mathbb{R}^{H \times W \times 5} \) and binary occupancy masks \( \mathbf{B} \in \{0, 1\}^{H \times W \times 5} \).

\boldparagraph{Object primitives} We represent objects using 11 primitive categories: vegetation cuboid (\( \mathcal{VC} \)), vegetation ellipsoid (\( \mathcal{VE} \)), vehicle big (\( \mathcal{VB} \)), vehicle small (\( \mathcal{VS} \)), two-wheelers (\( \mathcal{TW} \)), human (\( \mathcal{H} \)), construction big (\( \mathcal{CB} \)), construction small (\( \mathcal{CS} \)), pole (\( \mathcal{P} \)), traffic control (\( \mathcal{TC} \)), and object (\( \mathcal{O} \)). All are annotated with 3D cuboids except \( \mathcal{VE} \), which uses 3D ellipsoids. Each primitive is parameterized by a 9D feature vector \( \mathbf{f}_{i} \in [0, 1]^9 \), comprising a normalized 3D center location \( \mathbf{t}_{i} \in \mathbb{R}^{3}\) and 6D Cholesky parameters \( \mathbf{c}_{i} \in \mathbb{R}^{6} \). The latter jointly encode orientation and size via the Cholesky decomposition~\cite{Cholesky1924} of a positive-definite scatter matrix representing the object's 3D extent. Unlike eigen- or quaternion-based encodings, this unique parameterization eliminates orientation ambiguities caused by rotational symmetries and improves training stability. Further details, including semantic class-to-primitive mappings and Cholesky derivations, are provided in the supplementary material.

\noindent Overall, we represent %
\( \mathcal{S} \) as \( \boldsymbol{\mathcal{L}} = \{\mathbf{H}, \mathbf{B}, \mathbf{F} \} \), where \( \mathbf{F} \in \mathbb{R}^{M \times 9} \) stores the feature vectors of all \( M \) object primitives.

\subsection{Layout Variational Autoencoder}
\label{subsec:lvae}
To enable LDM-based generation, we design a layout VAE (LVAE) with jointly trained raster and vector branches that encode 3D scene layouts into a compact 2D latent space.

\boldparagraph{Encoding} Our architecture employs two encoders for distinct input modalities. The convolutional \textbf{ground encoder}~\cite{Rombach2022CVPR} \( \mathcal{E}_{\mathcal{G}} \) encodes normalized ground raster maps to a latent code \( \mathbf{z}_{\mathcal{G}} \in \mathbb{R}^{h \times w \times c} \), where \( h = H/2^{d} \), \( w = W/2^{d} \), and \( c \) is the latent channel dimension. The \textbf{object encoder} \( \mathcal{E}_{\mathcal{O}} \) processes object primitives represented by feature vectors \( \mathbf{f}_{i} \). To support batching during training and fixed-size set prediction at decoding, we pad each layout to a constant number of primitives per category, determined as a conservative upper bound based on dataset statistics (see supplementary material). Each primitive is thus augmented with a 10D learnable category embedding and a binary padding flag \( p_i \in \{0, 1\} \), yielding features \( \boldsymbol{\mathcal{F}} \in \mathbb{R}^{N \times 20} \) for a fixed number \( N \) of objects across all categories. These features are projected to a higher-dimensional space and processed by a Transformer encoder~\cite{Vaswani2017NIPS} to model inter-primitive relations. The resulting embeddings are mapped onto a learnable 2D latent grid matching the ground latent resolution via a scatter-mean operation. Specifically, each feature is assigned to a grid cell based on the normalized 2D center location of its corresponding primitive, and overlapping entries are averaged. 
This design preserves permutation invariance, maintains spatial structure in the latent space, and empirically worked best in our initial experiments. %
The final grid serves as the object latent \( \mathbf{z}_{\mathcal{O}} \in \mathbb{R}^{h \times w \times c} \), which is channel-wise concatenated with \( \mathbf{z}_{\mathcal{G}} \) to form the joint layout latent \( \mathbf{z}_{\mathcal{L}} = [\mathbf{z}_{\mathcal{G}};\mathbf{z}_{\mathcal{O}}] \in \mathbb{R}^{h \times w \times 2c} \) for diffusion modeling.

\boldparagraph{Decoding} During decoding, a latent code is sampled from the joint distribution and split along the channel dimension into ground and object components. The \textbf{ground decoder}  \( \mathcal{D}_{\mathcal{G}} \) reconstructs raster maps from \( \mathbf{z}_{\mathcal{G}} \) using a convolutional architecture~\cite{Rombach2022CVPR}. For object primitives, the \textbf{object decoder} \( \mathcal{D}_{\mathcal{O}} \) is based on the DETR~\cite{Carion2020ECCV} Transformer. The latent \( \mathbf{z}_{\mathcal{O}} \) is patchified into \( h' \times w'\) tokens, which are linearly projected and augmented with positional embeddings. These serve as keys and values in the decoder's cross-attention layers. For each category, \( \mathcal{D}_{\mathcal{O}} \) uses a fixed number of learnable \textit{object queries}, yielding \( \boldsymbol{Q} \in \mathbb{R}^{N \times d_L} \), which are transformed into output embeddings. These are grouped by category and passed through category-specific Feed Forward Networks (FFNs) to predict 3D center locations \( \mathbf{\hat{t}}_{i}\), Cholesky parameters \( \mathbf{\hat{c}}_{i} \), and existence probabilities \( \hat{p}_{i} \) for all primitives. %

\boldparagraph{Training} Our first-stage training objective combines raster and vector losses with a KL divergence term over the joint latent (see supplementary material for a detailed formulation). For \textbf{ground raster maps}, we apply an L1 loss on height maps, averaged over occupied pixels, and a binary cross-entropy (BCE) loss on occupancy masks. For \textbf{object primitives}, predicted and ground-truth instances are first matched within each category via the Hungarian algorithm~\cite{Kuhn1955Hungarian}, minimizing a pairwise cost consisting of (i) BCE loss on existence probabilities, (ii) L1 loss on center locations, and (iii) L1 loss on Cholesky parameters. After matching, the corresponding losses are computed per category, normalized by the number of real ground-truth instances in the batch, and then averaged across the batch.

\boldparagraph{Layout Reconstruction} We recover the 3D layout by fusing per-class height maps into a single height field, which is extruded into a triangle mesh. Object primitives with existence probability above a threshold are placed using their predicted centers and Cholesky parameters. This enables handling both sparse and dense scenes while maintaining a fixed-size prediction set. As with the ground layout, object primitives are meshed for visualization. Additional implementation details are provided in the supplementary material.

\subsection{Diffusion Transformer and Applications}
\label{subsec:dit}
The second stage trains a LDM~\cite{Rombach2022CVPR} on the joint LVAE latent space. A RePaint-inspired~\cite{Lugmayr2022CVPR} latent manipulation strategy enables further downstream applications without fine-tuning.

\boldparagraph{Training} Joint layout latent codes %
are extracted from the frozen LVAE and used to train a diffusion model \( \epsilon_{\theta}(\mathbf{z}_{\mathcal{L}}^{t}, t, y) \), where \( y \) is a scene label controlling vegetation density (low, medium, or high). Although we focus on vegetation, this conditioning generalizes to other object categories and their combinations (see supplementary material). Following~\cite{Peebles2023ICCV}, we adopt a Transformer backbone %
and incorporate \( y \) via adaLN-Zero blocks~\cite{Perez2018AAAI}. The model is trained to predict Gaussian noise using a mean squared error (MSE) loss. %

\boldparagraph{Scene Inpainting} RePaint~\cite{Lugmayr2022CVPR} introduces a modified diffusion sampling scheme for inpainting that synchronizes the denoising of unknown pixels with the noised version of known pixels. We apply this strategy in the 2D latent space of our pre-trained DiT to enable 3D layout editing. A binary latent mask specifies editable regions, allowing diverse edits such as inpainting the left half of a scene while keeping the right side intact (see \figref{fig:teaser} (3a)). Our disentangled latent structure naturally supports ground-conditioned object generation, where only object channels are inpainted while ground channels remain fixed (see \figref{fig:teaser} (3b)).

\boldparagraph{Scene Outpainting} We extend a 3D layout block beyond its original spatial extent using a sliding-window approach that leverages the same manipulation strategy as inpainting. To preserve spatial consistency, each newly generated block is designed to partially overlap by 50\% with its adjacent block. The scene is first expanded in parallel along the four cardinal directions, followed by a second pass to fill the corner regions. This process can be iteratively applied to produce arbitrarily large 3D scenes.

\noindent Additional details on the diffusion objective and its applications are provided in the supplementary material.

\section{Experiments}
\label{sec:experiments}

\boldparagraph{Dataset} Although our method can be trained on any dataset with 3D primitive annotations for arbitrary semantic categories, we focus on the underexplored urban setting and conduct our main experiments on KITTI-360~\cite{Liao2022PAMI}. KITTI-360 annotations cover entire urban environments rather than only selected object classes, resulting in many primitives per scene and a demanding evaluation setting. We remove rare labels and group the remaining ones by semantic and scale similarity into 16 categories. Each layout covers a \(64\,m \times 64\,m\) area centered on the ego vehicle. To minimize spatial overlap between training and test data, we thoroughly split the dataset and discard frames with incomplete annotations, yielding 61,913 training and 1,233 test poses. Split details and additional results on Argoverse 2~\cite{Argoverse22021NeurIPS} are provided in the supplementary material.

\begin{table}[t]
\centering
\footnotesize
\resizebox{\columnwidth}{!}{%
\begin{tabular}{lcccr}
\toprule
\textbf{Method} & \textbf{Resolution} & \textbf{Size} (MB) $\downarrow$ &
\textbf{IoU}$\uparrow$ & \textbf{mIoU}$\uparrow$ \\
\midrule
\multirow{3}{*}{SemCity~\citep{Lee2024CVPR}}
 & $64^2\times8$ & 0.12 & 82.19 & 50.52 \\
 & $128^2\times16$ & 1.00 & 90.38 & 68.98 \\
 & $256^2\times32$ & \third{8.00} & 81.75 & \third{70.52} \\
\midrule
 SemCity-1M & $256^2\times32$ & 8.00 & \second{97.69} & \best{93.81} \\
\midrule
\multirow{2}{*}{XCube~\citep{Ren2024CVPR}}
 & $64^2\times8$ & 0.051 & 82.76 & 35.52 \\
 & $256^2\times32$ & \second{3.05} & \best{99.97} & \second{79.47} \\
\midrule
PrITTI (voxelized) & $256^2\times32$ & \best{2.52} & \third{90.58} & 70.27 \\
\bottomrule
\end{tabular}%
}
\vspace{-0.3cm}
\caption{\textbf{Stage 1: Voxel reconstruction results} comparing IoU, mIoU, and mean per-sample memory for voxel-based baselines and voxelized PrITTI. Despite being natively primitive-based (and hence at a disadvantage), PrITTI remains competitive on these metrics. \bestcap{Best}, \secondcap{second-best}, and \thirdcap{third-best} results refer to the same (finest) resolution, where native size comparisons are meaningful.}
\label{tab:reconstruction_quantitative}
\end{table}

\begin{figure}[h]
\centering
\includegraphics[width=\linewidth]{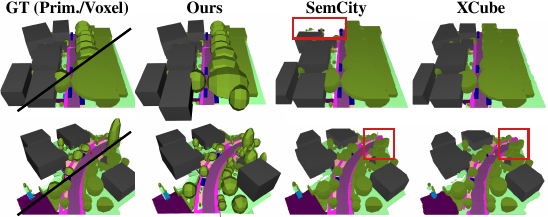}
\vspace{-0.4cm}
\caption{\textbf{Stage 1: Qualitative reconstruction results} on the same test scenes shown in each method’s native representation: primitives for PrITTI and voxel grids for SemCity and XCube. Voxel-based methods sometimes yield incomplete geometry and grid-induced distortions, such as vertical clipping at tall primitives.} 
\label{fig:reconstruction_qualitative}
\end{figure}

\boldparagraph{Baselines} As the first primitive-based method for urban 3D semantic scene generation, we compare against recent voxel-based baselines: SemCity~\cite{Lee2024CVPR}, PDD~\cite{Liu2024ECCV}, and XCube~\cite{Ren2024CVPR}. %
SemCity employs a two-stage pipeline, training a triplane autoencoder (AE) %
followed by a triplane diffusion model. We adopt its setup and voxelize our primitive scenes into a \( 256^{2} \times 32 \) grid. %
PDD~\cite{Liu2024ECCV} and XCube~\cite{Ren2024CVPR} are hierarchical voxel methods that refine scenes across scales. PDD applies discrete diffusion over a three-level pyramid 
(\( 32^{2} \times 4 \), \( 64^{2} \times 8 \), \( 256^{2} \times 16 \)), while XCube uses a VAE-diffusion paradigm across two hierarchy levels (\( 64^{2} \times 8 \), \( 256^{2} \times 32 \)).

\boldparagraph{Implementation Details} We implement the raster AE and LDM using standard VAE~\cite{Rombach2022CVPR} and DiT~\cite{Peebles2023ICCV} backbones from Diffusers~\cite{diffusers}. Ground raster maps have resolution \( H=W=256 \), and during encoding, we use a downsampling factor \( d=8\) and latent channel dimension \( c=32\). Both the object encoder and decoder use six Transformer layers, the latter processing \( N=514 \) object queries. 
The diffusion model is based on the DiT-B backbone %
with DDPM~\cite{Ho2020NeurIPS} noise scheduling. %
Additional details are provided in the supplementary material. Next, we evaluate scene reconstruction (\secref{sec:stage_1}), followed by the full generative model (\secref{sec:stage_2}).

\begin{table}[t!]
  \centering
  \footnotesize
  \setlength{\tabcolsep}{4pt}
  \renewcommand{\arraystretch}{1.12}
  \begin{tabular}{l|cc|cc}
    \toprule
    \multirow{2}{*}{\textbf{Method}} & \multicolumn{2}{c|}{\textbf{Raster} \textcolor{gray}{($\times 10^{-2}$)}} & \multicolumn{2}{c}{\textbf{Primitive}} \\
    & \textbf{MSE} $\downarrow$ & \textbf{IoU} $\uparrow$ & \textbf{AP3D} $\uparrow$ & \textbf{AP3D@50} $\uparrow$ \\
    \midrule
    Default          & \second{0.75} & \best{99.96} & \best{62.12} & \best{46.96} \\
    w/o latent split & \third{3.55} & \second{99.70} & \third{53.78}          & \third{39.09}          \\
    w/o objects      & \best{0.59} & \best{99.96} & \textemdash & \textemdash \\
    w/o ground       & \textemdash & \textemdash & \second{60.28} & \second{46.37} \\
    \bottomrule
  \end{tabular}
  \vspace{-0.3cm}
  \caption{\textbf{LVAE ablations} for latent split and joint training.}
  \label{tab:ablation_quantitative}
\end{table}

\begin{figure}[t!]
  \centering
  \includegraphics[width=\linewidth,  keepaspectratio]{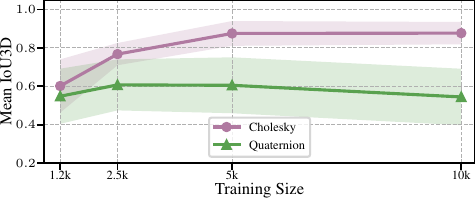}
  \vspace{-0.65cm}
  \caption{\textbf{Cholesky vs. quaternion} encodings across training sizes.}
  \label{fig:ablation_qualitative}
  \vspace{-0.6cm}
\end{figure}

\subsection{Stage 1: 3D Semantic Scene Reconstruction}
\label{sec:stage_1}

\begin{figure*}[t!]
\vspace{-0.0cm}
\centering
\includegraphics[width=\textwidth]{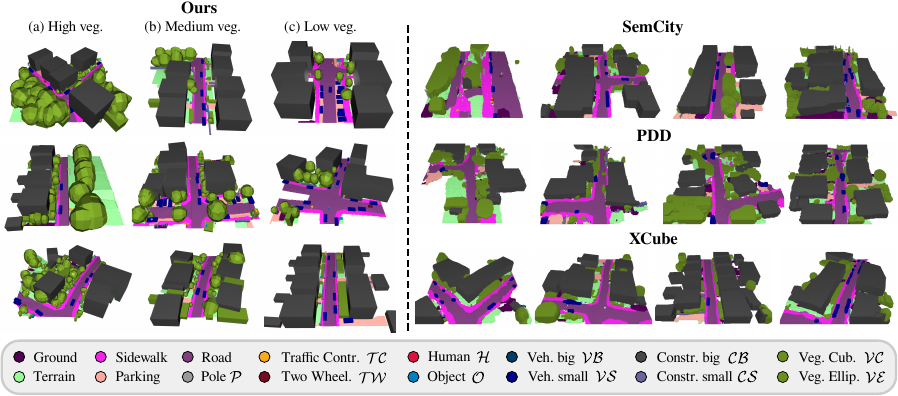} \\
\vspace{-0.25cm}
\caption{\textbf{3D Semantic Scene Generation.} Comparison of 3D semantic layouts generated by PrITTI (Ours, left) and voxel-based baselines (SemCity, PDD, XCube, right). PrITTI enables controllable generation, here conditioned on vegetation density (low, medium, or high), and produces more realistic, well-shaped scenes with clearer object boundaries. All baseline samples are generated unconditionally.}
\label{fig:03_generation}
\vspace{-0.3cm}
\end{figure*}

\boldparagraph{Metrics} To evaluate \textbf{3D scene reconstruction}, we compute standard voxel-based metrics~\cite{Lee2024CVPR}, including Intersection-over-Union (IoU) and mean IoU (mIoU). IoU is evaluated on a binary occupancy grid (foreground vs. background), while mIoU is averaged over all classes excluding the empty one. As PDD lacks a VAE component, we report results for SemCity, XCube, and our voxelized primitive reconstructions. Note that PrITTI is inherently primitive-based and hence at a disadvantage in this evaluation. However, it also supports %
standard 3D object detection metrics~\cite{Brazil2023CVPR}. Specifically, we report the mean \( AP_{3D}\) across all object categories (excluding \( \mathcal{VE} \)) and over a range of IoU3D~\cite{Ravi2020ARXIV} thresholds \( \tau \in \{0.05, \ldots, 0.50\} \), along with \( AP_{3D}^{50} \) at \( \tau = 0.50 \). For ground raster maps, we measure IoU for occupancy masks and MSE for height maps reconstruction.

\boldparagraph{Primitives vs. Voxels} Our method combines ground raster maps with primitives, yielding an edit-friendly but also more compact and memory-efficient representation than voxel-based baselines. As shown in \tabref{tab:reconstruction_quantitative}, it requires only 2.52 MB per scene (2.50 MB for rasters, 0.02 MB for primitives) versus 8.00 MB for SemCity and 3.05 MB for XCube at their finest resolution. By default, during AE training, SemCity’s decoder is evaluated on 400K sampled query points. We found that this %
limits performance on our voxelized dataset that preserves full volumetric geometry (\eg entire building volumes instead of outlines). We hence also evaluated an improved version with 1M query points, which greatly improved reconstruction (mIoU 70.52 $\rightarrow$ 93.81). We refer to this configuration as SemCity-1M. To reduce memory usage, we also train our baselines at lower resolutions and upsample the outputs to \( 256^2 \times 32\). This reduces per-scene memory requirements but causes a sharp drop in reconstruction quality (see \tabref{tab:reconstruction_quantitative}).
While upsampled models are more memory-efficient and sometimes comparable to ours (\eg SemCity at \( 128^{2} \times 16\)), they cannot scale to higher resolutions, where finer grids demand prohibitive memory or incur severe upsampling artifacts. 
This highlights a clear trade-off between memory requirements and reconstruction quality in voxel-based approaches. In contrast, our representation is resolution-independent and achieves comparable reconstruction quality with constant, low memory, despite voxelization of our scenes introducing additional errors not present in voxel-native baselines.
Importantly, as will be demonstrated in \secref{sec:stage_2}, our primitive-based model exhibits superior generation capabilities, benefiting from its compact and semantically-structured 3D scene representation.

\boldparagraph{Ablations} We analyze autoencoder design and object parameterization variants in \tabref{tab:ablation_quantitative} and \figref{fig:ablation_qualitative}: \textbf{(i) Cholesky vs. quaternions:} We compare our Cholesky-based encoding to a quaternion-based alternative on synthetic samples, each containing one zero-centered vehicle primitive with random scale and yaw. We vary the training set size while keeping the test set fixed across all settings (see supplementary material for details). As shown in \figref{fig:ablation_qualitative}, our parameterization consistently achieves higher mean IoU3D and greater stability, while the quaternion baseline plateaus and even degrades with more data, likely due to sign ambiguity causing loss discontinuities. \textbf{(ii) Latent split:} In this ablation (see \tabref{tab:ablation_quantitative} ``w/o latent split''), both decoders operate on the joint latent \( \mathbf{z}_{\mathcal{L}} \) instead of their respective components \( \mathbf{z}_{\mathcal{G}} \) and  \( \mathbf{z}_{\mathcal{O}} \). This modification degrades performance across all metrics (\eg AP3D: 62.12 $\rightarrow$ 53.78; MSE: 0.0075 $\rightarrow$ 0.0355), suggesting each decoder benefits from domain-specific features. \textbf{ (iii) Joint vs. separate training:}  We compare joint training of ground and object branches in a shared latent space against training them independently. In the default setup, a single latent is sampled from the joint distribution and split during decoding, whereas the ablated models (see \tabref{tab:ablation_quantitative} ``w/o objects'' and ``w/o ground'') use no shared latent or joint optimization. Although separate training slightly improves height map MSE (0.0059 vs. 0.0075), joint training achieves higher AP3D (62.12 vs. 60.28) by promoting better semantic alignment between branches, which enables context-aware object placement (\eg cars on road).

\subsection{Stage 2: 3D Semantic Scene Generation}
\label{sec:stage_2}

\boldparagraph{Metrics} We evaluate \textbf{3D scene generation} using Precision (Prec.)~\cite{Kynkaanniemi2019NeurIPS}, Recall (Rec.)~\cite{Kynkaanniemi2019NeurIPS}, Fréchet Inception Distance (FID)~\cite{Heusel2017NeurIPS}, and Inception Score (IS)~\cite{Salimans2016NeurIPS}, measuring fidelity and diversity. %
Following~\cite{Paschalidou2021NEURIPS, Tang2024CVPR}, both the reference and generated 3D scenes are rendered into \( 256^{2} \) top-down semantic maps. As standard practice~\cite{Ho2020NeurIPS, Dhariwal2021NeurIPS, Karras2019CVPR, Karras2021NeurIPS}, we use the training set for reference %
and generate an equal number of outputs using 250 DDPM sampling steps. For PrITTI, we balance the conditioning labels by sampling an equal number of scenes per class (low, medium, and high) without classifier-free guidance, while all baseline methods are unconditional. We found Prec. and Rec. to be sensitive to factors such as the evaluation sample size, the diversity of reference scenes, and the neighborhood size used for manifold estimation. In practice, we select 1K reference samples that are maximally spaced apart and use a neighborhood size of \( k=3 \). Additional evaluation settings and results are provided in the supplementary material.

\begin{table}[t]
  \centering
  \begin{adjustbox}{max width=\linewidth}
    \begin{tabular}{l|l|cccc|c}
      \toprule
        \textbf{Method} & \textbf{Variant} & 
        \textbf{Prec.} $\uparrow$ & \textbf{Rec.} $\uparrow$ 
        & \textbf{FID} $\downarrow$ & \textbf{IS} $\uparrow$ & \textbf{Time} (s) $\downarrow$ \\
        \midrule
        SemCity~\citep{Lee2024CVPR} & 1M &  
        0.133 & 0.063 & 177.623 & 3.454 & 2.35\\
        \midrule
        \multirow{3}{*}{PDD~\citep{Liu2024ECCV}} 
        & Level 1 & 0.000 & 0.000 & 302.256 & 1.558 & \second{0.94}\\
        & Level 2 & 0.182 & 0.000 & 166.086 & 1.896 & 1.89\\
        & Level 3 & 0.156 & 0.021 & 159.878 & 2.711 & 12.84\\
        \midrule
        \multirow{2}{*}{XCube~\citep{Ren2024CVPR}} 
        & Level 1 & 0.243 & 0.008 & 140.455 & 2.397 & \third{1.41}\\
        & Level 2 & 0.482 & 0.230 & 94.822 & 3.480 & 3.50\\
        \midrule
        \multirow{3}{*}{PrITTI (Ours)} 
        & DiT-B  & \third{0.712} & \third{0.491} & \third{73.952} & \third{3.856} & \best{0.58}\\
        & DiT-L  & \second{0.807} & \second{0.545} & \best{71.028} & \best{4.210} & 1.78\\
        & DiT-XL & \best{0.812} & \best{0.616} & \second{71.735} & \second{3.964} & 2.11\\
        \bottomrule
    \end{tabular}
  \end{adjustbox}
  \vspace{-0.3cm}
  \caption{\textbf{Generation Results.} Comparison of PrITTI and baselines across generative metrics and mean generation time per scene.}
  \label{tab:generative}
  \vspace{-0.65cm}
\end{table}

\begin{figure*}[t!]
\vspace{-0.0cm}
\centering
\includegraphics[width=\textwidth]{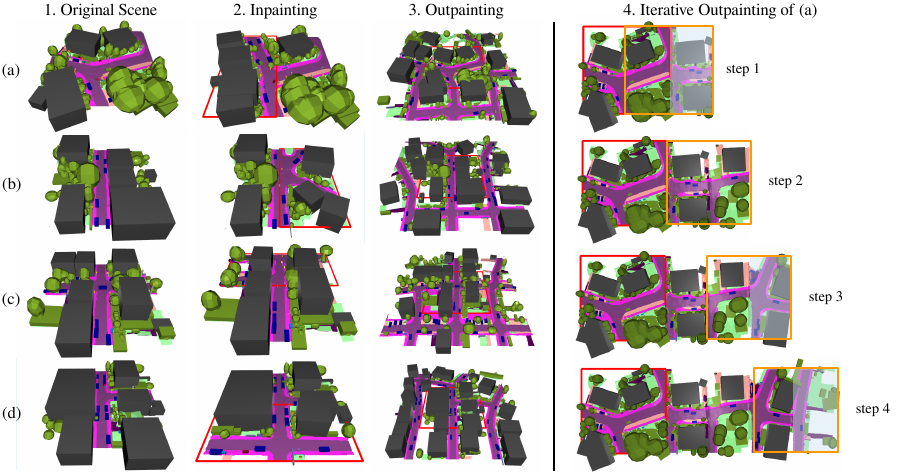} \\
\vspace{-0.3cm}
\caption{\textbf{Scene-level Editing.} (Left) Inpainting and outpainting results. Red outlines indicate the inpainted region (second column) or the original scene area (third column). (Right) An iterative outpainting example where the scene is progressively extended to the right. The red outline highlights the original scene boundary, while the orange denotes the current sliding-window region being generated at each step.}
\label{fig:04_applications}
\vspace{-0.5cm}
\end{figure*}

\boldparagraph{Results} For fair comparison, we compute all generative metrics using the same reference samples across methods. For hierarchical baselines, each level is evaluated on the same set of generated scenes, where 
outputs from coarser levels are upsampled to match the finest resolution. As shown in \tabref{tab:generative}, PrITTI significantly outperforms all baselines. We further evaluate larger diffusion model variants, DiT-L and DiT-XL, trained with the same configuration and total compute as DiT-B. Scaling to DiT-L yields notable improvements, while DiT-XL offers marginal gains. Regarding SemCity, we observe a large gap between reconstruction and generation quality, partly due to class-biased query point sampling during training and reconstruction versus uniform sampling during generation. This leads to overprediction of frequent classes seen in training, thereby reducing precision and recall. Moreover, its triplane-based diffusion design poses challenges in maintaining semantic consistency across planes, whereas our unified 2D latent space effectively mitigates these issues. Similarly, PDD shows low generative metrics across hierarchy levels. Coarser stages produce overly simplified outputs lacking fine structural detail, while the final level introduces additional noise and semantic inconsistencies, even after two weeks of training the corresponding diffusion model, as in the original setup. %
Furthermore, the split-scene generation strategy used at this level, where four $128^2 \times 16$ blocks are synthesized to form a $256^2 \times 16$ scene, may further complicate learning global consistency. Among the baselines, XCube attains the best generative performance. Qualitative results further highlight PrITTI's superior generation quality over baselines.
As shown in \figref{fig:03_generation}, PrITTI 
produces diverse and coherent scenes with well-shaped, clearly separated objects, while also supporting controllable generation. In contrast, baseline methods often yield irregular shapes, indistinct object boundaries, and semantically inconsistent regions. Their scenes generally lack fine structural detail and occasionally appear fragmented, with disjoint areas or missing surface regions.
Additional 3D and BEV-rendered results for all methods are included in the supplementary material, along with representative examples of low-quality generations that help contextualize the quantitative metrics.

\tabref{tab:generative} also reports the mean generation time per scene, averaged over 100 samples for all methods using the same number of diffusion timesteps. For hierarchical approaches, the reported time is cumulative across levels. Consequently, PDD and XCube exhibit longer generation times (12.84 seconds and 3.5 seconds, respectively) due to their multi-stage processing. Overall, PrITTI achieves the fastest generation (0.58), outperforming all baselines, including SemCity (2.35), which may be attributed to its more compact latent representation compared to SemCity’s triplane-based design. %

\subsection{Downstream Applications}

\boldparagraph{Scene-level Editing} %
As illustrated in \figref{fig:04_applications}, our method supports both inpainting and outpainting of 3D layouts. Inpainting results show that localized edits are achieved while preserving global structure, whereas outpainting results demonstrate that scenes are extended with diverse road layouts and coherently placed primitives. Our sliding-window mechanism (\figref{fig:04_applications}, right) progressively expands the scene by overlapping with the previously generated block, maintaining overlapping regions while synthesizing new content in the non-overlapping space. By iteratively applying this process, our model can generate large-scale, diverse, and semantically coherent urban layouts without additional training.

\boldparagraph{Object-level Editing} Our instance-level primitive representation enables intuitive object-wise editing of 3D scenes through direct geometric operations. As shown in \figref{fig:teaser}, primitives can be easily modified, such as being rotated or removed. Unlike voxel grids, which lack explicit object instances and require costly re-generation for simple edits, our approach supports direct manipulation by updating the corresponding primitive parameters (\eg center for translation). %

\boldparagraph{Photo-realistic Street-view Generation}
Given a generated 3D scene, we extract the road centerline from the binary road mask, sample camera viewpoints along it, and render 2D semantic maps. These maps are then used to condition a ControlNet~\cite{Zhang2023ICCVb} fine-tuned on KITTI-360~\cite{Lu2024ARXIV}. As shown in \figref{fig:teaser}, this enables realistic view synthesis consistent with the input semantics. Despite the coarse primitive geometry, our layouts effectively guide image generation, producing diverse object shapes beyond simple cuboids and ellipsoids.

\noindent Additional results for all applications, including large-scale scene extrapolation, are provided in the supplementary.

\section{Conclusion}
\label{sec:conclusion}
We introduced PrITTI, the first framework for large-scale 3D semantic urban scene generation using vectorized object primitives and rasterized ground surfaces. Our latent diffusion model generates high-quality, diverse, and controllable layouts while further supporting downstream tasks without fine-tuning. Compared to voxel-based baselines, PrITTI achieves similar reconstruction quality with lower memory requirements and stronger generative performance. In addition, unlike voxel grids, its instance-level primitive representation enables intuitive object-level edits. %

\boldparagraph{Limitations} %
In future work, we plan to address appearance synthesis 
with task-specific rendering pipelines tailored to our layouts~\cite{Lu2024ARXIV}. Learning more expressive primitives or deformable shapes promises to enhance geometric and visual fidelity~\cite{Aygun2024CVPR,Paschalidou2021CVPR}. Finally,
extending our approach to dynamic environments will pave the way toward autonomous driving simulations and diverse adversarial scenario generation~\cite{Hanselmann2022ECCV}.

\clearpage

\section*{Acknowledgments}
Andreas Geiger was supported by the ERC Starting Grant LEGO-3D (850533) and the DFG EXC number 2064/1 - project number 390727645.
Christina Ourania Tze and Daniel Dauner were supported by the German Federal Ministry for Economic Affairs and Energy within the project NXT GEN AI METHODS (19A23014S). We thank the International Max Planck Research School for Intelligent Systems (IMPRS-IS) for supporting Daniel Dauner. Christina Ourania Tze acknowledges her membership in the European Laboratory for Learning and Intelligent Systems (ELLIS) PhD program.

{
    \small
    \bibliographystyle{ieeenat_fullname}
    \bibliography{bib/bibliography_long,bib/bibliography,bib/bibliography_custom}
}

\clearpage

\maketitlesupplementary

\setcounter{section}{0}
\renewcommand\thesection{\Alph{section}}
\renewcommand\thesubsection{\Alph{section}.\arabic{subsection}}

In this supplementary material, we provide additional details on dataset preprocessing, implementation, and experimental setup. We also present further quantitative results evaluating generative performance and additional qualitative examples for all applications discussed in the main paper. Our \href{https://youtu.be/mk6wO3U5aB8}{supplementary video} offers a brief motivation and overview of our method as well as extended visual results.

\section{Dataset Details}
In this section, we describe our dataset construction and the key procedures used to analyze it.

\boldparagraph{Dataset Construction} KITTI-360~\cite{Liao2022PAMI} defines 37 labels, which we merge into 16 semantic categories, covering all major urban scene components. As an initial step, we analyze the dataset statistics and remove a few extremely rare labels including tunnel, bridge, train, stop, and caravan. Among the remaining ones, five correspond to ground-related classes (road, terrain, sidewalk, parking, and ground), which we treat as distinct ground categories.
The rest are grouped into 11 object categories based on both semantic and physical scale similarity, the latter being important for accurate primitive reconstruction. A complete mapping of semantic classes to object categories is provided in \tabref{tab:object_categories}. For detailed definitions of individual classes, we refer readers to~\cite{Liao2022PAMI}.

For each ego-vehicle pose, we adopt its IMU coordinate system as the local reference frame and define a \( 64\,m \times 64\,m\) 3D layout centered on the vehicle. The field of view (FOV) extends \( 64\,m\) 
in the forward direction and \( 32\,m\) laterally on each side. To avoid including distant or marginally visible objects, we retain only primitives whose 3D centers lie within this FOV and ground polygons with at least one vertex contained in it. Poses without any primitive annotations (typically at sequence boundaries) are excluded. The provided KITTI-360 test poses are sequentially distributed over limited areas of each sequence, covering only a narrow spatial extent. In short sequences, they typically span one continuous region, while in longer sequences, they may occupy two disjoint areas. Consequently, the original split insufficiently represents the full spatial diversity of the scenes.

To address this, we construct a custom train-test split. We first compute the total driving distance across all KITTI-360 sequences and divide the full trajectory into 100 equal-distance segments. Within each segment, the majority of poses are assigned to the training set, while a \(15\,m\) region near the end is reserved for testing. To minimize spatial overlap with training poses, we exclude \(5\,m\) margins on either side of the test regions. Furthermore, we remove any test pose located within \(5\,m\) in the Bird's Eye View (BEV) of a training pose, and vice versa. This prevents overlap caused by the ego-vehicle revisiting the same location from opposite directions. The final split comprises 61,913 training poses and 1,233 test poses, ensuring broad and non-overlapping spatial coverage across the entire dataset.

\begin{table*}[htbp]
\centering
\small
\renewcommand{\arraystretch}{1.2}
\begin{tabular}{l|lll}
\toprule
\textbf{Object Category} & \textbf{Semantic Classes} & \textbf{Primitive Type} & \textbf{Predicted Count}\\
\midrule
vegetation cuboid $(\mathcal{VC})$ & vegetation & 3D cuboid & 178\\
vegetation ellipsoid $(\mathcal{VE})$ & vegetation & 3D ellipsoid & 159\\
vehicle big $(\mathcal{VB})$ & truck, bus & 3D cuboid & 2\\
vehicle small $(\mathcal{VS})$ & car, trailer, unknown vehicle & 3D cuboid & 18\\
two-wheelers $(\mathcal{TW})$ & motorcycle, bicycle & 3D cuboid & 6\\
human $(\mathcal{H})$ & rider, pedestrian & 3D cuboid & 5\\
construction big $(\mathcal{CB})$ & building, garage, unknown construction & 3D cuboid & 16\\
construction small $(\mathcal{CS})$ & wall, fence, guardrail, gate & 3D cuboid & 77\\
pole $(\mathcal{P})$ & small pole, big pole & 3D cuboid & 19\\
traffic control $(\mathcal{TC})$ & traffic light, traffic sign, lamp & 3D cuboid & 17\\
object $(\mathcal{O})$ & trashbin, box, unknown object & 3D cuboid & 17\\
\bottomrule
\end{tabular}
\vspace{-0.2cm}
\caption{List of object categories defined in our method along with their corresponding semantic classes and annotated primitive types. The last column indicates the fixed number of primitives predicted per category, determined based on dataset statistics.}
\label{tab:object_categories}
\end{table*}

\boldparagraph{Primitive Count Selection} We adopt a fixed-size set prediction strategy for object primitives rather than an autoregressive decoding scheme, as this offers clear advantages for large-scale 3D scenes. Predicting all primitives in parallel enables the model to handle hundreds of objects in a single forward pass, whereas an autoregressive decoder would require sequential sampling, leading to higher inference time and cumulative prediction errors. In addition, generating the entire scene concurrently avoids imposing an arbitrary ordering on objects, which is inconsistent with the inherently unordered nature of scene layouts. Accordingly, we design the object decoder \( \mathcal{D}_{\mathcal{O}}\) (see \secref{subsec:lvae_supp}) to infer a fixed-size set of \( N^c\) primitives for each category \( c\). We determine \( N^c\) by analyzing the distribution of ground-truth primitive counts and selecting the 95\textsuperscript{th} percentile as the cutoff. The resulting values for each category are listed in \tabref{tab:object_categories}. During preprocessing, if the count of instances in a category exceeds its predefined threshold in a given scene, we retain only the \(N^c\) closest objects to the ego-vehicle and discard the farthest ones. This ensures that all scenes, regardless of their density, are processed. Importantly, not all decoded objects are necessarily retained in the final scene. Each predicted primitive is associated with an existence probability, and only those exceeding a predefined threshold are kept. This probabilistic filtering allows the model to naturally handle both sparse and dense scenes while maintaining a fixed-size set. 

\boldparagraph{Scene Labeling} To train our class-conditional latent diffusion model (see \secref{subsec:dit_supp}), each scene is assigned a discrete label reflecting its vegetation density. Since such annotations are not provided in the KITTI-360 dataset, we derive them automatically from dataset statistics. Specifically, for each scene, we compute the total number and volume of vegetation primitives, considering both cuboids and ellipsoids. A scene is then labeled as \textit{low} if both quantities fall below their respective 25\textsuperscript{th} percentile thresholds, \textit{high} if both exceed the 75\textsuperscript{th} percentile, and \textit{medium} otherwise.

\section{Methodological Details}
This section describes the representations, architectural components, and implementation details of our method.

\subsection{Cholesky Parameters}
\label{subsec:cholesky_supp}
Standard quaternion-based encodings of 3D orientation suffer from sign ambiguity: both \( \mathbf{q} \) and \( -\mathbf{q} \) represent the same rotation but differ numerically. This causes discontinuities in the loss landscape when optimized with standard regression objectives. Similarly, eigendecomposition-based encodings are ambiguous because eigenvectors are only defined up to sign (\ie both 
\( \mathbf{v} \) and \( -\mathbf{v} \)
are valid but numerically distinct), leading to comparable instability during gradient-based training. To overcome these issues, we propose a Cholesky-based parameterization that jointly encodes orientation and size in a continuous 6D representation. 

\boldparagraph{Scatter Matrix Construction} To derive this representation, we first define a scatter matrix \( \mathbf{S} \in \mathbb{R}^{3 \times 3} \) that describes the spatial extent of an object in 3D space. Let \( \lambda_1, \lambda_2, \lambda_3 > 0 \) denote the object's scaling factors along its principal axes, and let \( \mathbf{v}_1, \mathbf{v}_2, \mathbf{v}_3 \in \mathbb{R}^3 \) be the orthonormal column vectors of its rotation matrix. Defining the orthonormal basis matrix \( \mathbf{V} = [\mathbf{v}_1\ \mathbf{v}_2\ \mathbf{v}_3] \) and the diagonal scale matrix \( \mathbf{\Lambda} = \mathrm{diag}(\lambda_1, \lambda_2, \lambda_3) \), the scatter matrix is given by:

\begin{equation}
    \mathbf{S} = \mathbf{V} \mathbf{\Lambda} \mathbf{V}^\top = \sum_{j=1}^{3} \lambda_j \mathbf{v}_j \mathbf{v}_j^\top
\end{equation}

\boldparagraph{Cholesky Decomposition} We apply the Cholesky decomposition~\cite{Cholesky1924} to the scatter matrix \( \mathbf{S} \), yielding 
\begin{equation}
    \mathbf{S} = \mathbf{L} \mathbf{L}^\top
\end{equation}

\noindent where \( \mathbf{L} \in \mathbb{R}^{3 \times 3} \) is a lower-triangular matrix. The Cholesky parameters \( \mathbf{c} \in \mathbb{R}^6 \) are defined as the six non-zero entries of \( \mathbf{L} \). This representation removes orientation ambiguities by operating on the sign-invariant scatter matrix \( \mathbf{S} \): since each outer product \( \mathbf{v}_j \mathbf{v}_j^\top\) is unchanged under sign flips \(\mathbf{v}_j \mapsto -\mathbf{v}_j\), \( \mathbf{S} \) remains invariant. 

\boldparagraph{Scatter Matrix Properties} Additionally, we can show that \( \mathbf{S} \) is symmetric and positive-definite: 

\noindent \textit{Symmetric}:
\begin{equation}
    \mathbf{S}^\top = (\mathbf{V} \mathbf{\Lambda} \mathbf{V}^\top)^\top =(\mathbf{V}^\top)^\top\mathbf{\Lambda}^\top\mathbf{V}^\top = \mathbf{V}\mathbf{\Lambda}\mathbf{V}^\top = \mathbf{S}
\end{equation}
since \( \mathbf{\Lambda} \) is diagonal and therefore symmetric.

\noindent \textit{Positive-definite}:  
Because \( \mathbf{V} \) is orthonormal (\ie \( \mathbf{V}^\top = \mathbf{V}^{-1}\)), \( \mathbf{S} \) is similar to \( \mathbf{\Lambda} \):
\begin{equation}
    \mathbf{S} = \mathbf{V} \mathbf{\Lambda} \mathbf{V}^{-1}
\end{equation}
and thus shares the same eigenvalues \( \lambda_1, \lambda_2, \lambda_3 > 0 \). %

\noindent These properties guarantee a unique Cholesky decomposition~\cite{Cholesky1924} with a lower-triangular matrix \( \mathbf{L} \) having strictly positive diagonal entries. This in turn defines a \textit{unique} 6D representation \( \mathbf{c} \in \mathbb{R}^6\), well-suited for regression-based learning.

\boldparagraph{Reconstruction} Given predicted Cholesky parameters \( \hat{\mathbf{c}}\), we reshape them into a lower-triangular matrix \( \hat{\mathbf{L}} \in \mathbb{R}^{3 \times 3}\) and recover the scatter matrix as \( \hat{\mathbf{S}} = \hat{\mathbf{L}} \hat{\mathbf{L}}^\top \). Performing eigendecomposition of \( \hat{\mathbf{S}} \) yields its eigenvalues \( \hat{\lambda}_{1}, \hat{\lambda}_{2}, \hat{\lambda}_{3}\) and eigenvectors \( \hat{\mathbf{v}}_1, \hat{\mathbf{v}}_{2}, \hat{\mathbf{v}}_{3}\). These define the object's scale matrix \( \hat{\mathbf{\Lambda}} = \mathrm{diag}(\hat{\lambda}_1, \hat{\lambda}_2, \hat{\lambda}_3)\) and rotation matrix \( \hat{\mathbf{R}} = [\hat{\mathbf{v}}_1\ \hat{\mathbf{v}}_2\ \hat{\mathbf{v}}_3] \).

\subsection{Layout Variational Autoencoder}
\label{subsec:lvae_supp}

In the first training stage, a Layout Variational Autoencoder (LVAE) learns to compress 3D semantic scene layouts into compact 2D latent representations suitable for subsequent latent diffusion modeling.

\subsubsection{Architecture}
The LVAE comprises separate encoder-decoder pairs for the ground and object modalities. 

\boldparagraph{Ground Encoder} Our ground encoder \( \mathcal{E}_{\mathcal{G}} \) consists of stacked downsampling blocks that progressively reduce spatial resolution while increasing channel dimensionality. Each block contains a residual unit followed by a strided convolution. Residual units include two convolutional layers with group normalization and SiLU activation. At the bottleneck, a self-attention layer captures global context, while additional residual blocks refine local features. The encoder takes as input the ground raster maps, \( [\mathbf{H};\mathbf{B}] \in \mathbb{R}^{256 \times 256 \times 10}\), and outputs a latent grid of shape \(32 \times 32 \times 64\). This grid is split along the channel dimension into two chunks representing the mean \( \boldsymbol{\mu}_{\mathcal{G}} \in \mathbb{R}^{32 \times 32 \times 32}\) and variance \( \boldsymbol{\sigma}_{\mathcal{G}} \in \mathbb{R}^{32 \times 32 \times 32}\) of the ground latent representation. 

\boldparagraph{Object Encoder} The object encoder \( \mathcal{E}_{\mathcal{O}} \) operates on a fixed-size set of object primitives, each represented by a 20D feature vector. These vectors are first projected to a 512D embedding space and then processed by a 6-layer Transformer encoder with 8 attention heads and a feedforward dimension of 1024. A subsequent linear projection layer reduces the output dimensionality to 64. The resulting per-object features are spatially arranged into a latent grid of shape \( 32 \times 32 \times 64\) using a scatter-mean operation. Similar to the ground branch, the grid is split along the channel dimension into the mean \( \boldsymbol{\mu}_{\mathcal{O}} \in \mathbb{R}^{32 \times 32 \times 32}\) and variance \( \boldsymbol{\sigma}_{\mathcal{O}} \in \mathbb{R}^{32 \times 32 \times 32}\) of the object latent representation.

\boldparagraph{Joint Layout Latent} We construct joint mean and variance tensors, \( \boldsymbol{\mu}_{\mathcal{L}}, \boldsymbol{\sigma}_{\mathcal{L}} \in \mathbb{R}^{32 \times 32 \times 64}\), by concatenating the corresponding ground and object components along the channel axis. A joint layout latent representation \( \mathbf{z}_{\mathcal{L}} \in \mathbb{R}^{32 \times 32 \times 64} \) is then sampled via the reparameterization trick. This formulation maintains a disentangled latent structure, enabling straightforward separation into ground and object latents, \( \mathbf{z}_{\mathcal{G}}, \mathbf{z}_{\mathcal{O}} \in \mathbb{R}^{32 \times 32 \times 32} \). We found this design beneficial for improving reconstruction fidelity and enabling further applications such as ground-conditioned primitive generation.

\boldparagraph{Ground Decoder} The ground decoder \( \mathcal{D}_{\mathcal{G}} \) processes the ground latent representation \( \mathbf{z}_{\mathcal{G}} \) through a self-attention layer followed by two residual units. The resulting features are progressively upsampled to the original spatial resolution of \( 256 \times 256\) using a cascade of upsampling blocks, each comprising two residual units and a transposed convolution. As the spatial resolution increases, the channel dimensionality is gradually reduced. A final convolutional layer maps the output to 10 channels, matching the dimensionality of the input ground raster maps, \([\hat{\mathbf{H}};\hat{\mathbf{B}}] \in \mathbb{R}^{256 \times 256 \times 10}\).

\boldparagraph{Object Decoder} The object decoder \( \mathcal{D}_{\mathcal{O}}\) adopts a DETR~\cite{Carion2020ECCV} Transformer architecture with \( N \) learnable queries representing the total number of predicted primitives across all object categories. It consists of 6 Transformer decoder layers with 8 attention heads, a feedforward dimension of 1024, and no dropout. The queries interact with the object latent \( \mathbf{z}_{\mathcal{O}} \) through cross-attention. Specifically, \( \mathbf{z}_{\mathcal{O}} \) is first patchified with a patch size of 1, producing a sequence of \( 32 \times 32 \) tokens augmented with sine positional embeddings. The resulting embeddings, each of dimension 512, are grouped by category and passed to category-specific prediction heads.

\boldparagraph{Prediction Heads} Each object category employs a dedicated 3-layer feedforward network with a hidden size of 1024. For every query, the head outputs a 9D vector encoding the normalized 3D center location and 6D Cholesky parameters of its corresponding primitive. An additional linear layer predicts its probability of existence.

\subsubsection{Training}
The first-stage training loss consists of three components, described below.

\boldparagraph{Ground Loss} We supervise the reconstruction of input raster maps using: (i) a binary cross-entropy (BCE) loss for occupancy masks and (ii) an L1 loss for height maps, averaged over occupied pixels. Let \( \mathbf{H} \in \mathbb{R}^{H \times W \times 5} \) and \( \mathbf{B} \in \{0,1\}^{H \times W \times 5} \) denote the ground-truth height maps and occupancy masks, and \( \hat{\mathbf{H}} \) and \( \hat{\mathbf{B}} \) their predicted counterparts. The ground loss is defined as:

\begin{equation}
\resizebox{0.9\hsize}{!}{$
\mathcal{L}_{\text{ground}} =
\lambda_{\mathcal{G}}^{\text{occ}} \cdot \mathcal{L}_{\text{BCE}}(\hat{\mathbf{B}}, \mathbf{B})
+ \lambda_{\mathcal{G}}^{\text{height}} \cdot
\frac{1}{\sum \mathbf{B}}
\big\| (\hat{\mathbf{H}} - \mathbf{H}) \odot \mathbf{B} \big\|_1
$}
\end{equation}

\noindent where \( \odot \) denotes element-wise multiplication. We set the loss weights \( \lambda_{\mathcal{G}}^{\text{occ.}} = 1 \) and \( \lambda_{\mathcal{G}}^{\text{height}} = 9\) to balance the relative contributions of the two terms.

\boldparagraph{Object Loss}  
The object decoder \( \mathcal{D}_{\mathcal{O}}\) predicts a fixed number of \( N^{c}\) primitives per object category \( c\) (see \tabref{tab:object_categories}), leading to \( N = \sum_{c=1}^{11} N^{c}\) total predictions per scene. Because the outputs are unordered, we establish one-to-one correspondences between predicted and ground-truth primitives using bipartite matching for each category independently. Specifically, let \(\mathcal{O}^{c} = \{\mathbf{o}_i \}_{i=1}^{N^c} \) and \( \hat{\mathcal{O}}^{c} = \{ \hat{\mathbf{o}}_j \}_{j=1}^{N^c} \) denote the sets of ground-truth and predicted primitives for category \( c \), respectively. Each primitive is parameterized as \( \mathbf{o}_{i} = (\mathbf{t}_{i}, \mathbf{c}_{i}, p_{i})\), where \( \mathbf{t}_{i} \in \mathbb{R}^{3}\) is its 3D center location, \( \mathbf{c}_{i} \in \mathbb{R}^{6}\) are the Cholesky parameters, and \( p_{i} \in \{0, 1\}\) distinguishes valid primitives \( (p_{i} = 1)\) from padded placeholders \( (p_{i} = 0)\). We apply the Hungarian algorithm~\cite{Kuhn1955Hungarian} to find the optimal assignment \( \hat{\sigma}^c \) minimizing the total matching cost:

\begin{equation}
   \hat{\sigma}^c = \arg \min_{\sigma^{c} \in \mathcal{P}_{N^c}} \sum_{i=1}^{N^c} \mathcal{L}_{\text{match}}(\mathbf{o}_i, \hat{\mathbf{o}}_{\sigma(i)})
   \label{eq:permutation}
\end{equation}

\noindent where \( \mathcal{P}_{N^c} \) is the set of all permutations of \( N^c \) elements. The pairwise matching cost is defined as:

\begin{equation}
\begin{aligned}
\mathcal{L}_{\text{match}}(\mathbf o_i,\hat{\mathbf o}_{\sigma(i)})
&= \lambda_{\mathcal O}^{\text{prob}} \,
   \mathcal{L}_{\text{BCE}}(p_i,\hat p_{\sigma(i)}) \\[2pt]
&\quad + \mathds{1}_{[p_i>0]} \,
   \lambda_{\mathcal O}^{\text{center}} \,
   \lVert \mathbf t_i - \hat{\mathbf t}_{\sigma(i)} \rVert_{1} \\[-2pt]
&\quad + \mathds{1}_{[p_i>0]} \,
   \lambda_{\mathcal O}^{\text{Chol}} \,
   \lVert \mathbf c_i - \hat{\mathbf c}_{\sigma(i)} \rVert_{1}
\end{aligned}
\label{eq:match-loss}
\end{equation}

\noindent After obtaining the optimal assignments, we compute the object loss over all categories as:

\begin{equation}
\mathcal{L}_{\text{object}} = \sum_{c \in \mathcal{C}} \left( \sum_{i=1}^{N^c} \mathcal{L}_{\text{match}}(\mathbf{o}_{i}, \hat{\mathbf{o}}_{\hat{\sigma}^c(i)}) \right)
\label{eq:line_loss}
\end{equation}

\noindent where \( \mathcal{C}\) denotes the set of object categories. Each term in~\eqnref{eq:line_loss} is normalized by the number of real (\ie non-padded) ground-truth primitives in the corresponding sample, and the normalized losses are averaged across the batch. The weights \( \lambda_{\mathcal{O}}^{\text{prob.}}, \lambda_{\mathcal{O}}^{\text{center}}\), and \( \lambda_{\mathcal{O}}^{\text{Chol.}} \) are configured differently for bipartite matching and for the final loss computation. For \textbf{matching}, we use category-agnostic constants: \( \lambda_{\mathcal{O}}^{\text{prob.}} = 6 \), \( \lambda_{\mathcal{O}}^{\text{center}} = 3 \), and \( \lambda_{\mathcal{O}}^{\text{Chol.}} = 3 \). This weighting emphasizes the BCE term to discourage pairing low-confidence predictions with true instances, while assigning equal importance to the center and Cholesky components for accurate geometric alignment. During \textbf{loss computation}, the weights vary across object categories according to their fixed number of predicted primitives. In particular, categories are grouped into low-, medium-, and high-count ranges, with higher weights assigned to higher-count groups. This adjustment compensates for normalization by the number of real primitives. Without it, high-count categories, typically containing more real instances per sample, would contribute less to the overall loss, leading to underfitting. Within each group, the center and Cholesky weights are balanced so that their corresponding losses have comparable magnitudes, ensuring that both geometric components contribute equally. The BCE term receives a smaller weight than the geometric terms, as existence confidence is already reinforced by the previous matching stage. Increasing its weight would disproportionately emphasize classification over geometry. The final group-specific weights are summarized in \tabref{tab:object_loss_weights}.

\begin{table}[h]
\centering
\resizebox{\columnwidth}{!}{
\begin{tabular}{l|l|ccc}
\toprule
\textbf{Group} & \textbf{Object Categories} & 
\(\lambda^{\text{prob.}}_{\mathcal{O}}\) & 
\(\lambda^{\text{center}}_{\mathcal{O}}\) & 
\(\lambda^{\text{Chol.}}_{\mathcal{O}}\) \\
\midrule
Low    & \(\mathcal{VB}, \mathcal{TW}, \mathcal{H}\) & 1 & 3  & 2  \\
Medium & \(\mathcal{VS}, \mathcal{CB}, \mathcal{P}, \mathcal{TC}, \mathcal{O}\) & 3 & 9  & 7  \\
High   & \(\mathcal{VC}, \mathcal{VE}, \mathcal{CS}\) & 5 & 15 & 12 \\
\bottomrule
\end{tabular}}
\vspace{-0.2cm}
\caption{Category-specific loss weights grouped by object count.}
\label{tab:object_loss_weights}
\end{table}

\boldparagraph{KL Loss} Following a standard VAE formulation~\cite{Kingma2014ICLR}, we apply a Kullback-Leibler (KL) divergence regularization on the joint layout latent. The loss is defined as:

\begin{equation}
\resizebox{0.9\hsize}{!}{$
\mathcal{L}_{\text{KL}}
= \lambda^{\text{KL}} \cdot
\left(
  -\frac{1}{2} \sum_{d=1}^{64}
  \big(
    1 + \log(\sigma_{\mathcal{L},d}^{2})
    - \mu_{\mathcal{L},d}^{2}
    - \sigma_{\mathcal{L},d}^{2}
  \big)
\right)
$}
\end{equation}

\noindent where \( \mu_{\mathcal{L},d} \) and \( \sigma_{\mathcal{L},d} \) denote the mean and standard deviation of the \( d\)-th latent dimension. We set \(\lambda^\textrm{KL}= 1\mathrm{e}^{-6}\).

\boldparagraph{Total Loss} The overall training objective for the first stage combines the three components as:

\begin{equation}
    \cL_\textrm{LVAE} =  \cL_\textrm{ground} + \cL_\textrm{object} + \cL_\textrm{KL}
\end{equation}

\subsubsection{Layout Reconstruction}

\begin{figure}[t]
  \centering
  \includegraphics[width=0.7\linewidth]{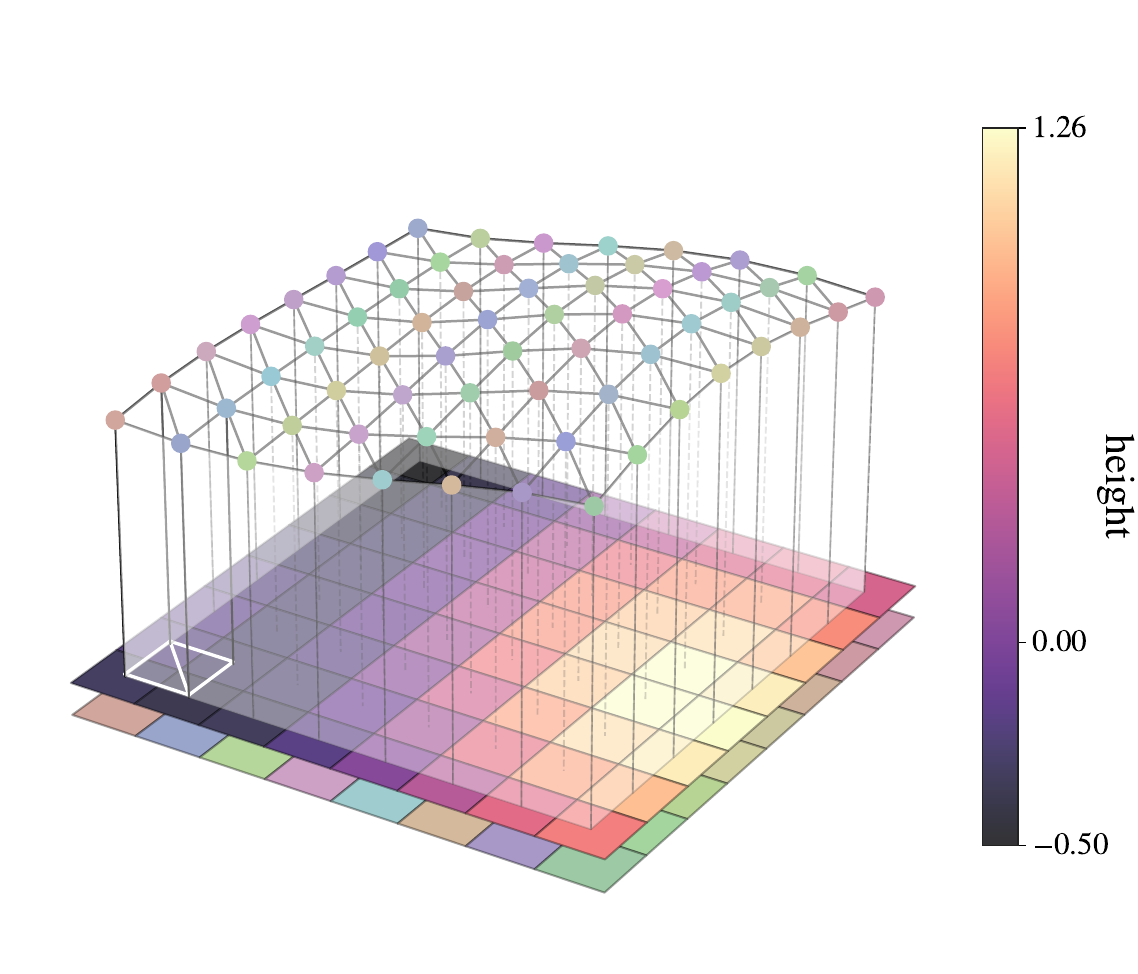}
  \caption{\textbf{Extrusion Process.}
  An illustrative extrusion example showing the composite color map (bottom), height field (middle), and lifted triangle mesh (top).}
  \vspace{-0.3cm}
  \label{fig:extrusion}
\end{figure}

To reconstruct a 3D scene layout, we follow separate procedures for ground surfaces and objects, reflecting their distinct representations. 
For \textbf{ground surfaces}, we first fuse the per-class height maps into a composite height field by selecting, at each pixel, the highest elevation and recording the class that contributed it. Using a fixed label-to-color mapping, this process simultaneously yields a composite color map aligned with this height field. The resulting raster representation is then converted into a triangle mesh (see \figref{fig:extrusion}). Each occupied pixel $(u,v)$ is lifted to a 3D vertex $(x,y,z)$ using its height value and the vertex color is assigned from the composite color map at the same location. To obtain a continuous surface, we triangulate each $2\times2$ pixel block at indices $(i,j), (i,j+1), (i+1,j),$ and $(i+1,j+1)$ using a consistent diagonal, forming two adjacent triangles per quad. For \textbf{objects}, we first derive the rotation matrix \( \hat{\mathbf{R}} \) and scale matrix \( \hat{\mathbf{\Lambda}} \) from the predicted Cholesky parameters \( \hat{\mathbf{c}} \in \mathbb{R}^6 \), as described in \secref{subsec:cholesky_supp}. We then construct the full transformation matrix \( \hat{\mathbf{T}} \in \mathbb{R}^{4 \times 4} \) by combining these with the predicted 3D center location \( \hat{\mathbf{t}} \in \mathbb{R}^3 \):

\begin{equation}
    \hat{\mathbf{T}} =
\begin{bmatrix}
\hat{\mathbf{R}} \hat{\mathbf{\Lambda}} & \hat{\mathbf{t}} \\
\mathbf{0}^\top & 1
\end{bmatrix}
\end{equation}

\noindent This transformation is applied to a unit-sized, zero-centered 3D cuboid or ellipsoid (depending on the object category) to position the primitive within the reconstructed scene layout.

\subsubsection{Implementation Details} The LVAE comprises 93.1M parameters and is trained using PyTorch Lightning\footnote{\url{https://lightning.ai/docs/pytorch/stable/}} for 3 days on a single A100 GPU with a batch size of 32. Optimization is performed using AdamW with a weight decay of \( 5e^{-4}\), combined with a OneCycleLR scheduler. The learning rate follows a cosine annealing schedule, starting at \( 1e^{-4}\) and decaying to \( 1e^{-5}\), without a warm-up phase. During inference, we discard object primitives whose predicted existence probability falls below a threshold of \( \hat{p} = 0.3\). We use a single global threshold across all semantic categories. This value controls the trade-off between retaining more primitives and suppressing low-confident ones: lower values lead to denser but potentially noisier layouts, while higher values yield cleaner but sparser reconstructions. In practice, we found that reconstruction quality is not highly sensitive to this parameter, and values within $[0.25,0.75]$ produce visually similar results.

\subsection{Diffusion Transformer}
\label{subsec:dit_supp}
In the second stage of our method, we train a latent diffusion model (LDM)~\cite{Rombach2022CVPR} on the joint layout latent space of our pre-trained LVAE.

\subsubsection{Architecture}
We adopt the standard DiT architectures from~\cite{Peebles2023ICCV} with a patch size of 2, yielding \( 16 \times 16\) tokens from the joint layout latent representation \( \mathbf{z}_{\mathcal{L}} \in \mathbb{R}^{32 \times 32 \times 64}\). \tabref{tab:dit_models} summarizes the configurations of the DiT Base (B), Large (L), and XLarge (XL) variants used in our experiments. 

\begin{table}[h]
  \centering
  \begin{adjustbox}{max width=\linewidth}
    \begin{tabular}{l|ccccc}
    \toprule
    \textbf{Model} & \textbf{Layers} & \textbf{Hidden Size} & \textbf{Heads} & \textbf{Batch Size} & \textbf{Params (M)} \\
    \midrule
    DiT-B  & 12 & 768  & 12 & 256 & 130 \\  
    DiT-L  & 24 & 1024 & 16 & 256 & 458 \\
    DiT-XL & 28 & 1152 & 16 & 256 & 675 \\
    \bottomrule
    \end{tabular}
  \end{adjustbox}
  \vspace{-0.2cm}
  \caption{Details of DiT-B, -L, and -XL model variants.}
  \label{tab:dit_models}
\end{table}

\subsubsection{Training} 
Given an initial latent \( \mathbf{z}_{\mathcal{L}}^0 \), we define a \textit{forward diffusion process} that gradually corrupts it by adding Gaussian noise over time, yielding a sequence of increasingly noisy latents \( \mathbf{z}_{\mathcal{L}}^1, \ldots, \mathbf{z}_{\mathcal{L}}^T \). The noise magnitude at each step is controlled by a variance schedule \( \{ \beta_t \in (0,1)\}_{t=1}^T \). The joint distribution %
over all timesteps is expressed as:
\begin{align}
    q(\mathbf{z}_{\mathcal{L}}^{1:T} \mid \mathbf{z}_{\mathcal{L}}^0) &= \prod_{t=1}^T q(\mathbf{z}_{\mathcal{L}}^t \mid \mathbf{z}_{\mathcal{L}}^{t-1}) \\
    q(\mathbf{z}_{\mathcal{L}}^t \mid \mathbf{z}_{\mathcal{L}}^{t-1}) &= \mathcal{N}(\mathbf{z}_{\mathcal{L}}^t; \sqrt{1 - \beta_t} \mathbf{z}_{\mathcal{L}}^{t-1}, \beta_t \mathbf{I})
\end{align}

\noindent A convenient property of the above process is that the noisy latent \( \mathbf{z}_{\mathcal{L}}^t \) at any timestep \( t \) can be sampled in a single step from the following distribution:
\begin{equation}
    q(\mathbf{z}_{\mathcal{L}}^t \mid \mathbf{z}_{\mathcal{L}}^0) = \mathcal{N}(\mathbf{z}_{\mathcal{L}}^t; \sqrt{\bar{\alpha}_t} \mathbf{z}_{\mathcal{L}}^0, (1 - \bar{\alpha}_t) \mathbf{I})
\end{equation}
where \( \alpha_t = 1 - \beta_t \), and \( \bar{\alpha}_t = \prod_{s=1}^t \alpha_s \). 

To synthesize new latents, we learn a \textit{reverse process} that denoises from pure Gaussian noise \( \mathbf{z}_{\mathcal{L}}^T \sim \mathcal{N}(0, \mathbf{I}) \) back to a clean latent \( \mathbf{z}_{\mathcal{L}}^0 \). The joint distribution of this process is:
\begin{align}
    p_\theta(\mathbf{z}_{\mathcal{L}}^{0:T}) = p(\mathbf{z}_{\mathcal{L}}^T) \prod_{t=1}^T p_\theta(\mathbf{z}_{\mathcal{L}}^{t-1} \mid \mathbf{z}_{\mathcal{L}}^t) \\
     p_\theta(\mathbf{z}_{\mathcal{L}}^{t-1} \mid \mathbf{z}_{\mathcal{L}}^t) = \mathcal{N}(\mathbf{z}_{\mathcal{L}}^{t-1}; \mu_\theta(\mathbf{z}_{\mathcal{L}}^t, t), \Sigma_\theta(\mathbf{z}_{\mathcal{L}}^t, t))
\end{align}

\noindent where each transition is modeled as a Gaussian with learnable mean \( \mu_\theta \) and covariance \( \Sigma_\theta \), estimated by a denoising network parameterized by \( \theta\). In practice, we follow~\cite{Ho2020NeurIPS} and set \( \Sigma_\theta(\mathbf{z}_{\mathcal{L}}^t, t) = \sigma_t^2 \mathbf{I} = \frac{1 - \bar{\alpha}_{t-1}}{1 - \bar{\alpha}_t} \beta_t \mathbf{I}\), treating it as an untrained, time-dependent constant. Moreover, we represent the mean \( \mu_\theta(\mathbf{z}_{\mathcal{L}}^t, t) \) as:
\begin{align}
    \mu_\theta(\mathbf{z}_{\mathcal{L}}^t, t) = \frac{1}{\sqrt{\alpha_t}} \left( \mathbf{z}_{\mathcal{L}}^t - \frac{\beta_t}{\sqrt{1 - \bar{\alpha}_t}} \, \epsilon_\theta(\mathbf{z}_{\mathcal{L}}^t, t) \right)
\end{align}
where \( \epsilon_\theta \) predicts the noise term \( \boldsymbol{\epsilon}\) in the reparameterization \( \mathbf{z}_{\mathcal{L}}^t = \sqrt{\bar{\alpha}_t} \, \mathbf{z}_{\mathcal{L}}^0 + \sqrt{1 - \bar{\alpha}_t} \, \boldsymbol{\epsilon} \). 

For \textit{training}, we adopt the simplified denoising objective of~\cite{Ho2020NeurIPS}:

\begin{equation}
\resizebox{0.8\hsize}{!}{$
    \mathcal{L}_{\text{simple}} =
    \mathbb{E}_{\mathbf{z}_{\mathcal{L}}^0, \boldsymbol{\epsilon}, t}
    \left[
        \left\|
            \boldsymbol{\epsilon} -
            \epsilon_\theta\!\left(
                \sqrt{\bar{\alpha}_t}\, \mathbf{z}_{\mathcal{L}}^0 +
                \sqrt{1 - \bar{\alpha}_t}\, \boldsymbol{\epsilon}, t
            \right)
        \right\|^2
    \right]
$}
\end{equation}

\noindent In our method, we further condition the denoising network on a scene label \(y\), controlling the density of specific scene primitives (\eg vegetation). The resulting conditional network \( \epsilon_\theta(\mathbf{z}_{\mathcal{L}}^t, t, y)\) is trained with the modified objective:

\begin{equation}
\resizebox{0.8\hsize}{!}{$
\mathcal{L}_{\text{LDM}} =
\mathbb{E}_{\mathbf{z}_{\mathcal{L}}^0, \boldsymbol{\epsilon}, t, y}
\left[
    \left\|
        \boldsymbol{\epsilon} -
        \epsilon_\theta\!\left(
            \sqrt{\bar{\alpha}_t}\, \mathbf{z}_{\mathcal{L}}^0 +
            \sqrt{1 - \bar{\alpha}_t}\, \boldsymbol{\epsilon},
            t, y
        \right)
    \right\|^2
\right]
$}
\end{equation}

\subsubsection{Applications}
Leveraging a latent manipulation strategy inspired by RePaint~\cite{Lugmayr2022CVPR}, our diffusion model enables practical downstream tasks without the need for additional fine-tuning.

\begin{figure*}[t!]
\centering
\includegraphics[width=\textwidth, height=0.3\textheight, keepaspectratio]{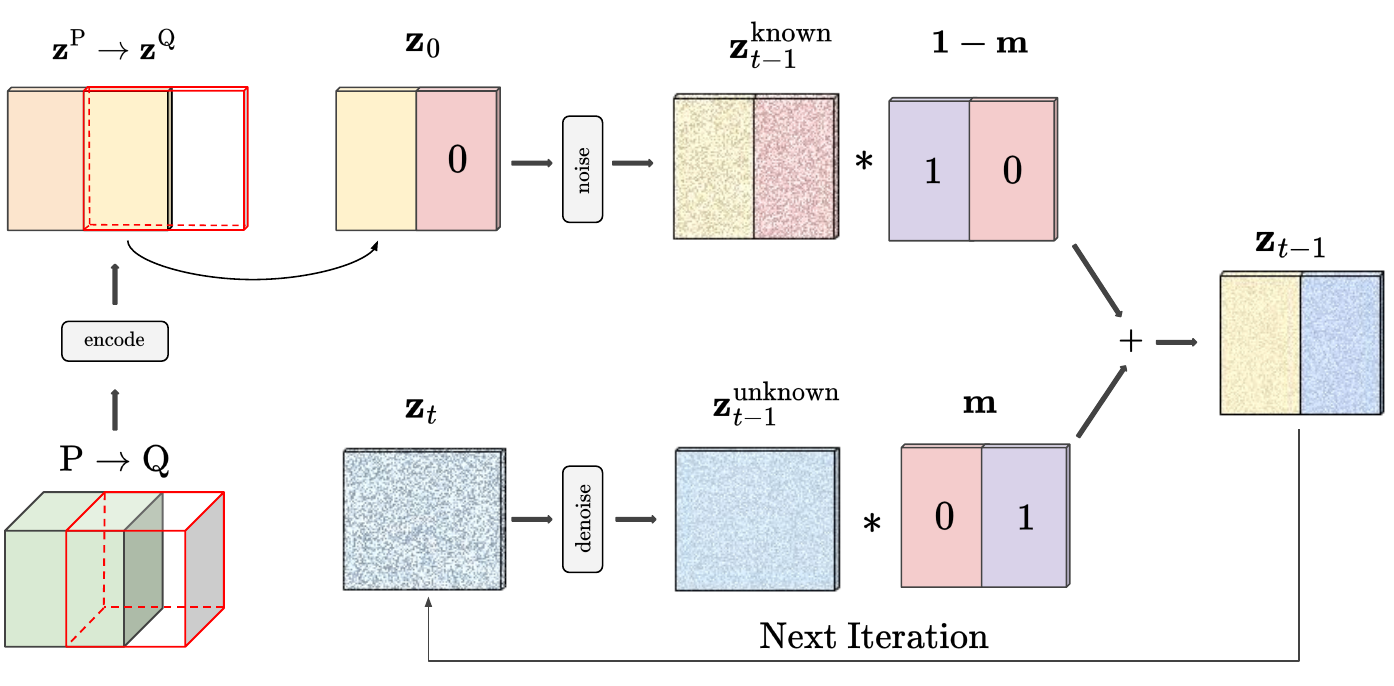} \\
\caption{\textbf{Latent-space Scene Outpainting.} Given a known layout block $P$ (green) with latent $\mathbf{z}^{P}$, the diffusion model predicts the latent $\mathbf{z}^{Q}$ for an adjacent block $Q$ (outlined in red) that partially overlaps with $P$. During each sampling step, the overlapping area (yellow) is re-noised to match the current diffusion timestep, while denoising is applied to $\mathbf{z}_{t}$. The resulting latents are then combined through the mask $\mathbf{m}$, yielding a coherent representation $\mathbf{z}_{t-1}$ for the next iteration that preserves the known content and refines the unknown region (blue).}
\label{fig:outpaitning_mechanism}
\vspace{-0.3cm}
\end{figure*}

\boldparagraph{Scene Inpainting} RePaint introduced a modified diffusion-based sampling strategy for image inpainting that employs a binary mask to distinguish known and unknown regions. The key idea is to leverage a pre-trained denoising diffusion model as a generative prior and to synchronize the denoising of unknown pixels with the noised version of known pixels. We adapt this approach to operate in the 2D latent space of a pre-trained LDM, enabling localized edits of 3D scene layouts. For simplicity, we omit the subscript \( \mathcal{L} \) and denote the noisy layout latent at timestep \( t\) as \( \mathbf{z}_{t} \). Given an initial latent \( \mathbf{z}_{0}\), the modified denoising update is formulated as:

\begin{align}
    \mathbf{z}_{t-1}^{\text{known}} &\sim \mathcal{N}(\sqrt{\bar{\alpha}_t} \mathbf{z}_{0}, (1 - \bar{\alpha}_{t})\mathbf{I}) \label{eq:z_known} \\
    \mathbf{z}_{t-1}^{\text{unknown}} &\sim \mathcal{N}(\mu_{\theta}(\mathbf{z}_{t}, t, y), \sigma_t^2 \mathbf{I}) \label{eq:z_unknown} \\
    \mathbf{z}_{t-1} &= (1 - \mathbf{m}) \odot \mathbf{z}_{t-1}^{\text{known}} + \mathbf{m} \odot \mathbf{z}_{t-1}^{\text{unknown}} \label{eq:z_composition}
\end{align}

\noindent where \( \mathbf{m} \) is a binary latent mask identifying the regions to be edited, \( \odot \) denotes element-wise multiplication, and \( \mathcal{N}(\mu_{\theta}(\mathbf{z}_{t}, t, y), \sigma_t^2 \mathbf{I}) \) represents the learned reverse process of our pre-trained diffusion model. By properly setting the mask \( m\), we can edit an original scene in diverse ways. 

\boldparagraph{Scene Outpainting}
Scene outpainting extends a 3D layout beyond its original spatial extent by predicting plausible continuations of existing structures. We perform this extrapolation directly in the latent space of our diffusion model, using the same manipulation strategy employed for inpainting. As illustrated in \figref{fig:outpaitning_mechanism}, the procedure adopts a sliding-window scheme: given a known layout block $P$ (green) with latent representation $\mathbf{z}^{P}$, we synthesize the latent $\mathbf{z}^{Q}$ of an adjacent block $Q$ (outlined in red) that overlaps with $P$. The goal is to generate $\mathbf{z}^{Q}$ such that the overlapping part is faithfully preserved while the extended area remains semantically and structurally aligned with $P$. In all experiments, we use an overlap of $50\%$. This choice balances structural coherence with generative freedom. A large value strengthens conditioning on the known region, promoting continuity of geometry and semantics across block boundaries, but it also restricts the model's ability to produce novel content. Conversely, overly reducing the overlap weakens the conditioning signal, often resulting in boundary artifacts and semantic misalignment. To initialize the diffusion process, we construct an input latent $\mathbf{z}_{0}$ by copying the overlapping region (yellow) from $\mathbf{z}^{P}$ and zeroing out the rest. A binary mask $\mathbf{m}$ marks the unknown area to be generated with ones and the known region with zeros. The process begins by sampling $\mathbf{z}_{T} \sim \mathcal{N}(\mathbf{0}, \mathbf{I})$, following the standard reverse diffusion formulation. During each step $t$, the latent $\mathbf{z}_{t}$ is jointly denoised across both regions, yielding an intermediate prediction $\mathbf{z}_{t-1}^{\text{unknown}}$ (see~\eqnref{eq:z_unknown}). To maintain statistical compatibility between the two parts, the forward noising process is applied to the input latent $\mathbf{z}_{0}$, yielding $\mathbf{z}_{t-1}^{\text{known}}$ (see~\eqnref{eq:z_known}). The two components are then merged through the mask $\mathbf{m}$ to form a unified latent representation $\mathbf{z}_{t-1}$ for the next denoising iteration (see~\eqnref{eq:z_composition}). 

To expand a scene spatially, we apply this mechanism in a two-stage approach. In the first pass, we generate the four cardinal neighbors (top, bottom, left, and right), each conditioned on its corresponding overlapping region with the original layout. In the second pass, we synthesize the remaining four corner regions (top-left, top-right, bottom-left, and bottom-right), which depend on the previously generated neighbors. We note that within each pass, all blocks are generated in parallel. This two-stage process effectively doubles the spatial extent of the original layout. Iteratively applying this expansion enables the synthesis of large-scale 3D scenes, a task we refer to as large-scale scene extrapolation. For simplicity, and consistent with prior work~\cite{Liu2024ECCV}, we restrict iterative extrapolation to a single spatial axis.

\subsubsection{Implementation Details}
All DiT variants are trained using the Diffusers library\footnote{\url{https://github.com/huggingface/diffusers}}. Training is performed for 3 days on 4 A100 GPUs with a batch size of 256. Optimization uses AdamW with a constant learning rate of \( 1e^{-4}\) and a weight decay of \( 1e^{-6}\). An Exponential Moving Average (EMA) with a decay rate of 0.9999 is applied to stabilize training. We employ the DDPM scheduler with 1000 timesteps and a linear variance schedule ranging from \(\beta_{\text{start}} = 0.0015\) to \(\beta_{\text{end}} = 0.015\). Classifier-free guidance~\cite{Ho2022ARXIV} is applied with a guidance scale of 4.0. During inference, we use 250 denoising steps, while inpainting and outpainting employ a roll-back of \( J = 10\) and \( R = 10\) resampling iterations.

\section{Experimental Details}

\begin{table*}[t]
\centering
\vspace{0.5em}
\resizebox{\linewidth}{!}{
\begin{tabular}{l|c|c|c|c|c|c}
\toprule
\textbf{Method} & \textbf{Hierarchical} & \textbf{Level} & \textbf{Voxel Size (m)} & \textbf{Voxel Resolution} & \textbf{VAE Train (days)} & \textbf{Diff. Train (days)} \\
\midrule
SemCity~\cite{Lee2024CVPR} & \xmark & Single & 0.25 & $256^2 \times 32$ & 3 & 3 \\
\midrule
\multirow{3}{*}{PDD~\cite{Liu2024ECCV}} & \multirow{3}{*}{\cmark} & Coarse & 2.0 & $32^2 \times 4$ & \textemdash & 1 \\
 &  & Medium & 1.0 & $64^2 \times 8$ & \textemdash & 2 \\
 &  & Fine & 0.25 & $256^2 \times 16$ & \textemdash & 14 \\
\midrule
\multirow{2}{*}{XCube~\cite{Ren2024CVPR}} & \multirow{2}{*}{\cmark} & Coarse & 1.0 & $64^2 \times 8$ & 1 & 2 \\
 &  & Fine & 0.25 & $256^2 \times 32$ & 6 & 7 \\
 \midrule
PrITTI (Ours) & \xmark & Single & \textemdash & \textemdash & 3 & 3 \\
\bottomrule
\end{tabular}
}
\caption{\textbf{Comparison of Architectures and Training Configurations.} For each method, we report the model hierarchy, resolution settings, and the corresponding VAE and diffusion training times (where applicable).}
\vspace{-0.3cm}
\label{tab:baselines}
\end{table*}

In this section, we describe the synthetic dataset and evaluation protocol used to compare our Cholesky-based encoding with a quaternion-based alternative. For this ablation study, we generate synthetic datasets of increasing size, where each sample contains a single zero-centered vehicle primitive with randomly sampled scales and yaw rotations. The training sets consist of 1.2K, 2.5K, 5K, and 10K samples, with each larger set including all samples from the smaller ones. A shared test set of 2K samples is used for evaluation. To reduce variance from random sampling, dataset generation is repeated with five different random seeds. For each seed and training size, both encoding variants are trained for 20 epochs, and the mean IoU3D is computed on the shared test set. Final results are reported as the average IoU3D over all seeds for each dataset size.

\section{Baseline Methods}
To the best of our knowledge, no prior work tackles 3D semantic urban layout generation using primitives. We therefore compare against three recent and well-established voxel-based methods, all trained from scratch on our voxelized dataset: SemCity~\cite{Lee2024CVPR}, PDD~\cite{Liu2024ECCV}, and XCube~\cite{Ren2024CVPR}. This comparison allows us to empirically demonstrate the benefits of primitives over voxels, the prevailing representation in this field, which is central to our contribution. SemCity adopts a continuous triplane diffusion model trained on the SemanticKITTI~\cite{Behley2019ICCV} and CarlaSC~\cite{Wilson2022RAL} datasets. It first trains a triplane autoencoder and then applies diffusion within the learned latent triplane space. The model follows a non-hierarchical design and supports downstream tasks such as semantic scene inpainting, outpainting, and completion refinement. PDD employs a discrete, hierarchical diffusion paradigm trained on the synthetic CarlaSC dataset. It generates scenes in a coarse-to-fine manner across three pyramid levels. The first level synthesizes scenes unconditionally from noise, while subsequent levels condition on the preceding outputs to progressively refine scene details. PDD supports large-scale scene outpainting and conditioned scene generation, the latter refining a ground-truth coarse layout into a more detailed semantic scene. XCube generates large-scale 3D scenes using a hierarchical sparse-voxel representation trained across the Karton City~\cite{KartonCity} and Waymo~\cite{Sun2020CVPR} datasets. Each level in the hierarchy is trained independently and consists of a VAE-diffusion pair trained sequentially (VAE first, then diffusion). Both the VAE and diffusion components use 3D UNet backbones. For urban scenes, XCube supports both unconditional and single-LiDAR-scan–conditioned generation, though the conditional implementation is not yet publicly available. \tabref{tab:baselines} summarizes the model and training configurations for all baseline methods and PrITTI. Following the original implementations, PDD and XCube are trained using 4 and 8 A100 GPUs, respectively. In addition to training configurations, we also compare the peak GPU memory usage of all methods under the same inference setup. As shown in~\tabref{tab:memory}, our method is more memory-efficient than voxel-based baselines, consistent with the use of a primitive-based representation.

\begin{table}[h]
    \centering
    \resizebox{\columnwidth}{!}{
    \begin{tabular}{l|c|c|c|c}
        \toprule
        \textbf{Metric} & \textbf{SemCity} & \textbf{PDD} & \textbf{XCube} & \textbf{Ours (DiT-B)} \\
        \midrule
        Peak GPU Memory (GB) & 8.30 & 1.33 & 4.84 & \textbf{0.89} \\
        \bottomrule
    \end{tabular}
    }
    \caption{Peak GPU memory usage (GB) under identical inference settings (100 scenes, batch size 1, single GPU).}
    \vspace{-0.5cm}
    \label{tab:memory}
\end{table}

\section{Additional Qualitative Results}
\label{sec:additional_results}
In this section, we present additional qualitative results for all tasks described in the main paper.

\begin{figure*}[t]
\centering
\includegraphics[width=\textwidth]{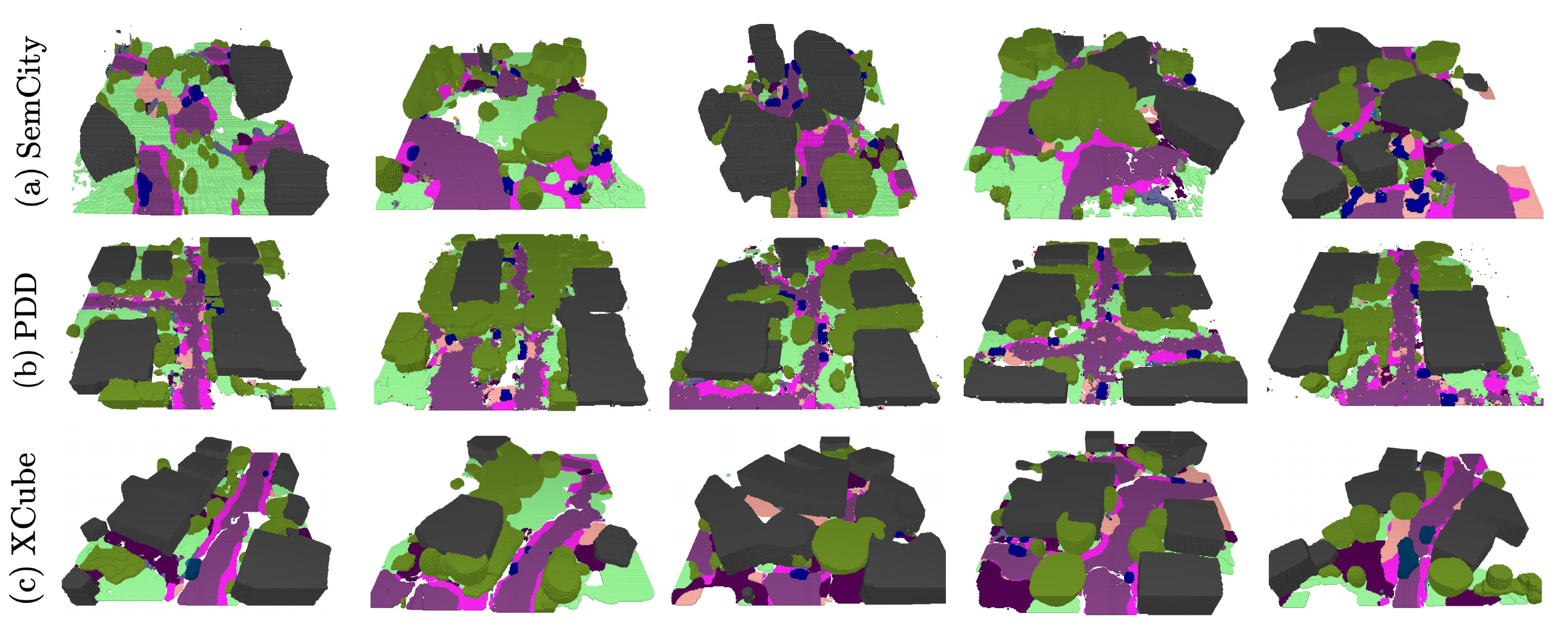} \\
\caption{\textbf{Low-quality scene generation examples} from baseline methods exhibiting fragmented geometry and semantic inconsistencies.}
\label{fig:baselines_failures}
\end{figure*}

\begin{figure*}[t]
\centering
\includegraphics[width=\textwidth,height=0.95\textheight, keepaspectratio]{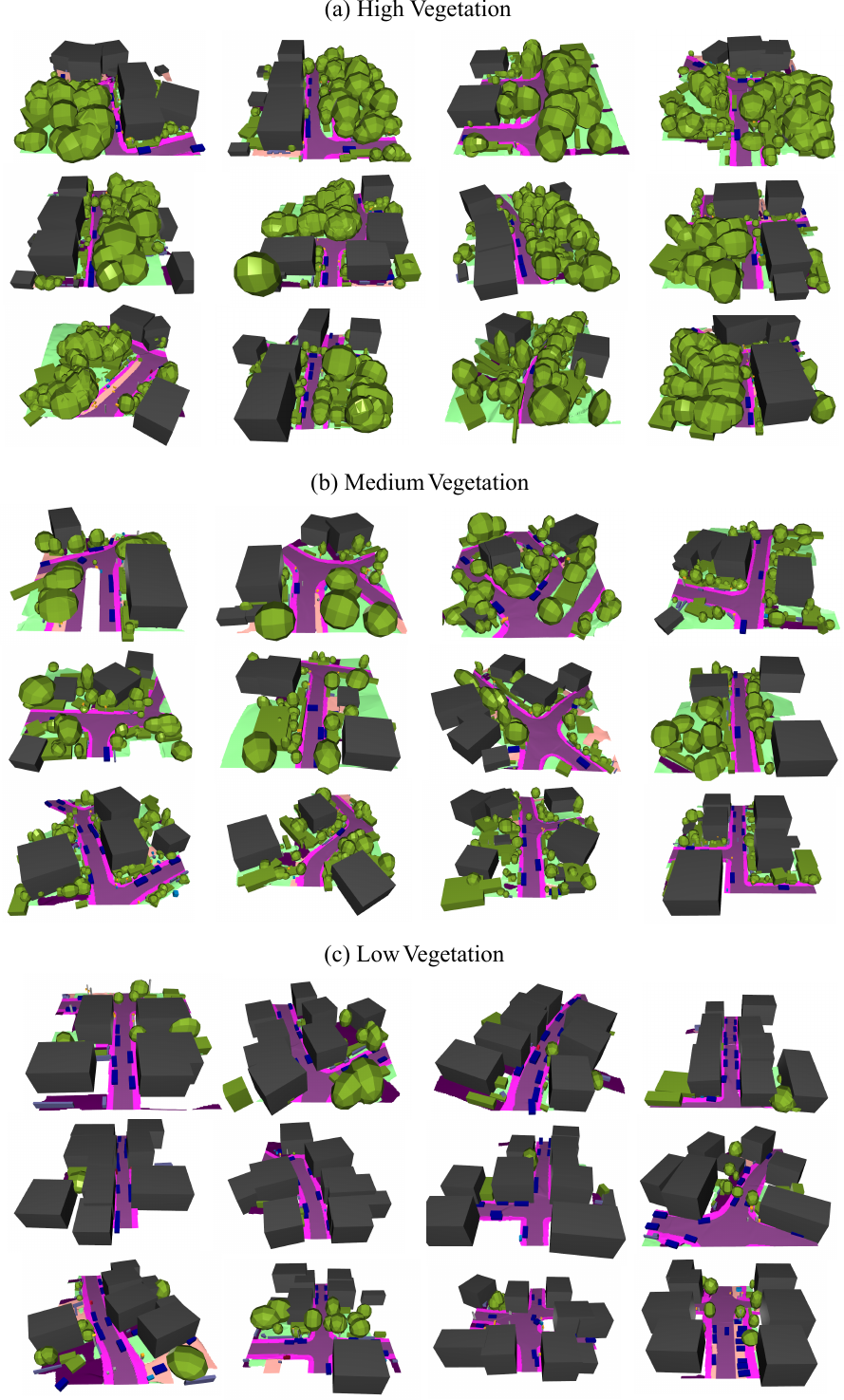}
\caption{\textbf{PrITTI Scene Generation Results.} Generated 3D semantic scenes grouped under three vegetation-density conditions: (a) high, (b) medium, and (c) low. Across all settings, PrITTI produces diverse and realistic urban layouts with coherent ground surfaces and well-shaped primitives placed in semantically appropriate locations.}
\label{fig:generation}
\end{figure*}

\begin{figure*}[t]
\centering
\includegraphics[width=\textwidth,height=0.95\textheight, keepaspectratio]{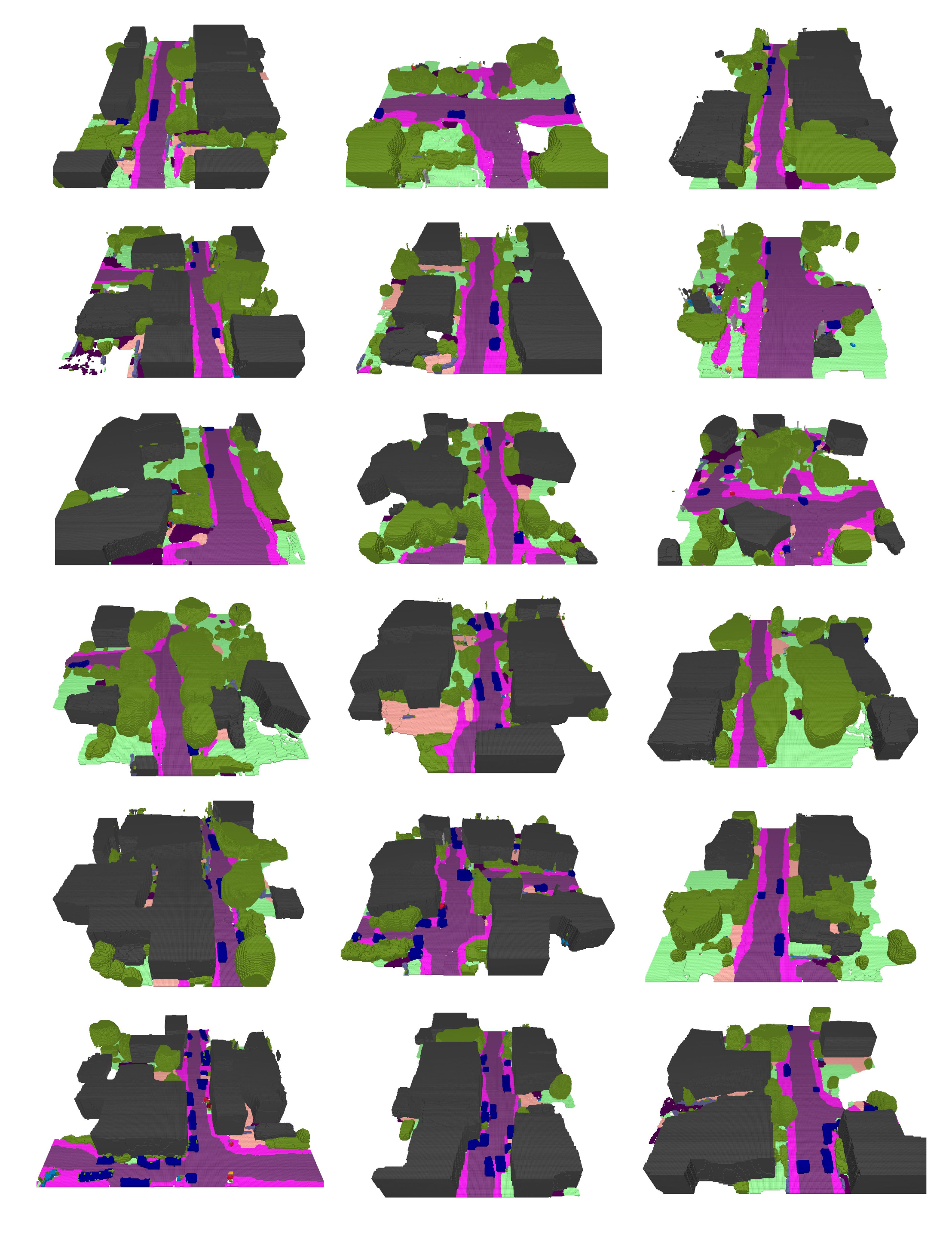}
\caption{\textbf{SemCity Scene Generation Results.} Unconditional 3D semantic scene samples generated by SemCity~\cite{Lee2024CVPR}, occasionally exhibiting irregular object shapes, blurred boundaries, and inconsistent semantic structures.}
\label{fig:semcity}
\end{figure*}

\begin{figure*}[t]
\centering
\includegraphics[width=\textwidth,height=0.95\textheight, keepaspectratio]{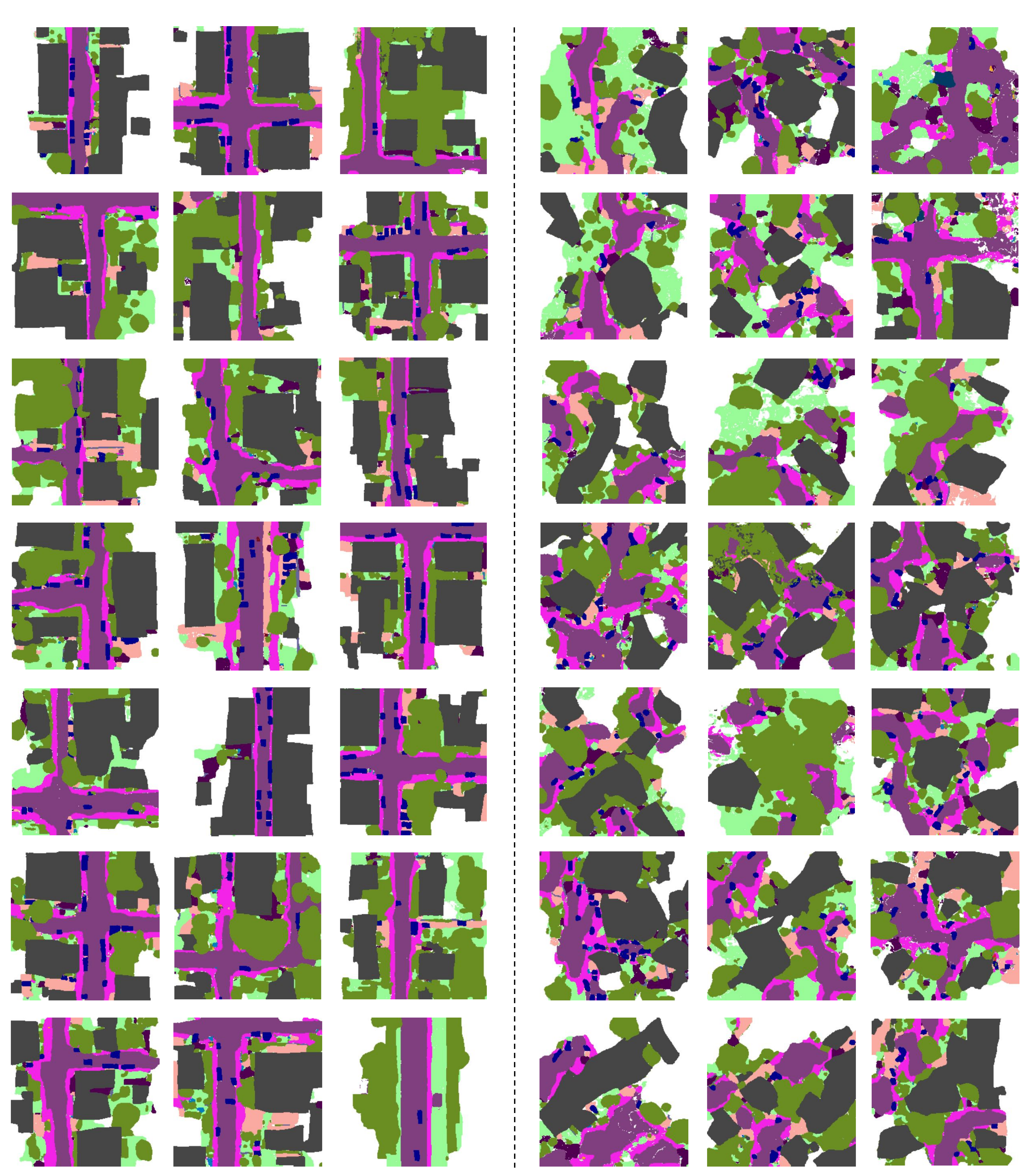}
\caption{\textbf{BEV Renderings of SemCity-generated Scenes.} The examples on the left show relatively well-structured and semantically coherent layouts, while those on the right exhibit fragmented, irregular geometry and inconsistent semantics.}
\label{fig:semcity_bev}
\end{figure*}

\begin{figure*}[t]
\centering
\includegraphics[width=\textwidth,height=0.95\textheight, keepaspectratio]{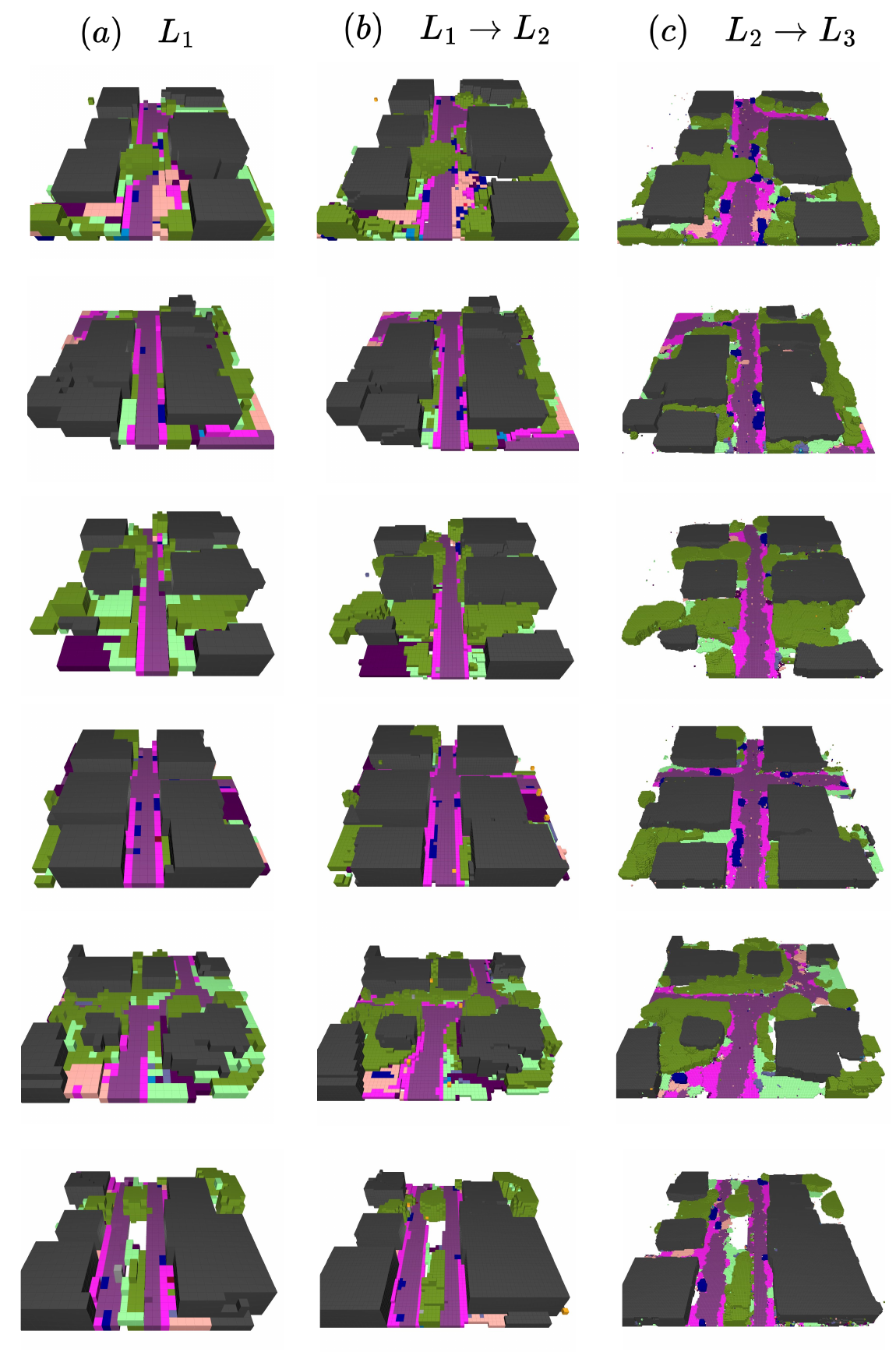}
\caption{\textbf{PDD Hierarchical Scene Generation Results.} Unconditional 3D semantic scenes generated by PDD~\cite{Liu2024ECCV} at successive pyramid levels. While scene details increase across levels, the final stage also introduces noticeable noise and semantic inconsistencies.}
\label{fig:pdd_hierarchical}
\end{figure*}

\begin{figure*}[t]
\centering
\includegraphics[width=\textwidth,height=0.95\textheight, keepaspectratio]{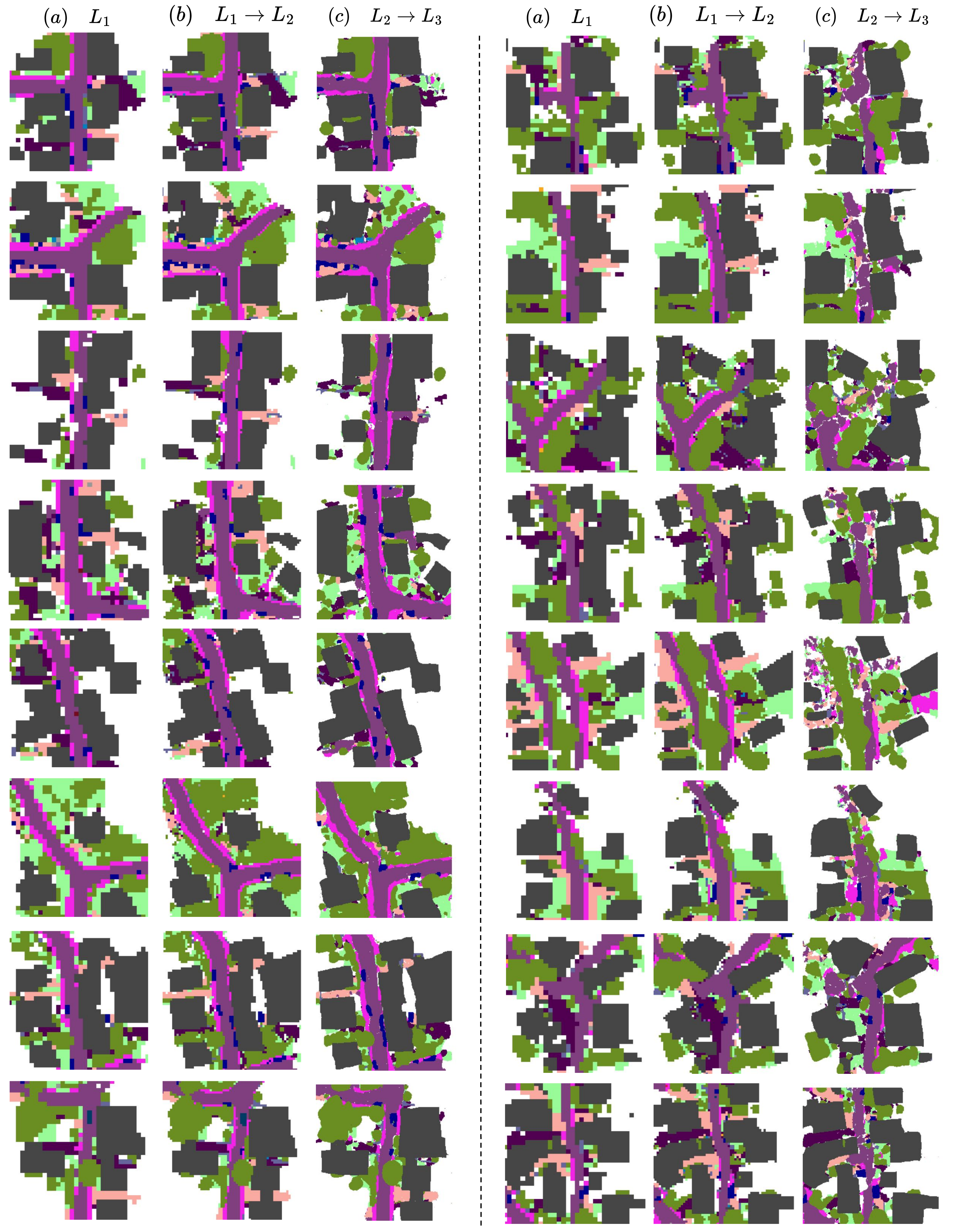}
\caption{\textbf{Hierarchical BEV Renderings of PDD-generated Scenes.} Each column triplet shows results across successive hierarchy levels. Examples on the left exhibit relatively coherent scene structure, whereas those on the right clearly show that the last refinement stage introduces noticeable noise and semantic inconsistencies, thereby reducing the perceived realism of the final outputs.}
\label{fig:pdd_bev}
\end{figure*}

\begin{figure*}[t]
\centering
\includegraphics[width=\textwidth,height=0.95\textheight, keepaspectratio]{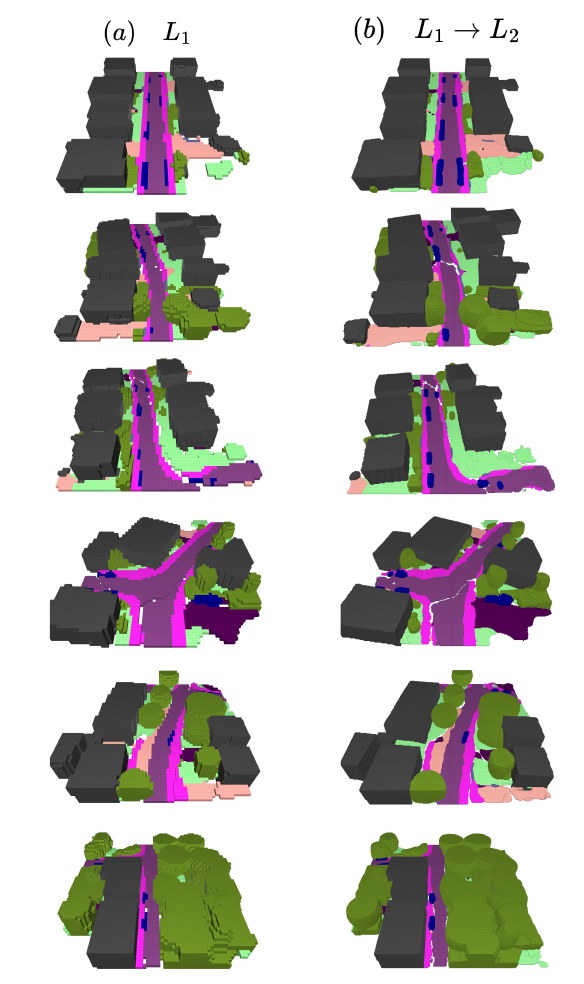}
\caption{\textbf{XCube Hierarchical Scene Generation Results.} Unconditional 3D semantic scenes generated by XCube~\cite{Ren2024CVPR} across successive hierarchy levels. Scene details improve across levels, and the overall structure appears realistic, but irregular object shapes (\eg vehicles) and occasional road-geometry discontinuities remain visible.}
\label{fig:xcube_hierarchical}
\end{figure*}

\begin{figure*}[t]
\centering
\includegraphics[width=\textwidth,height=0.95\textheight, keepaspectratio]{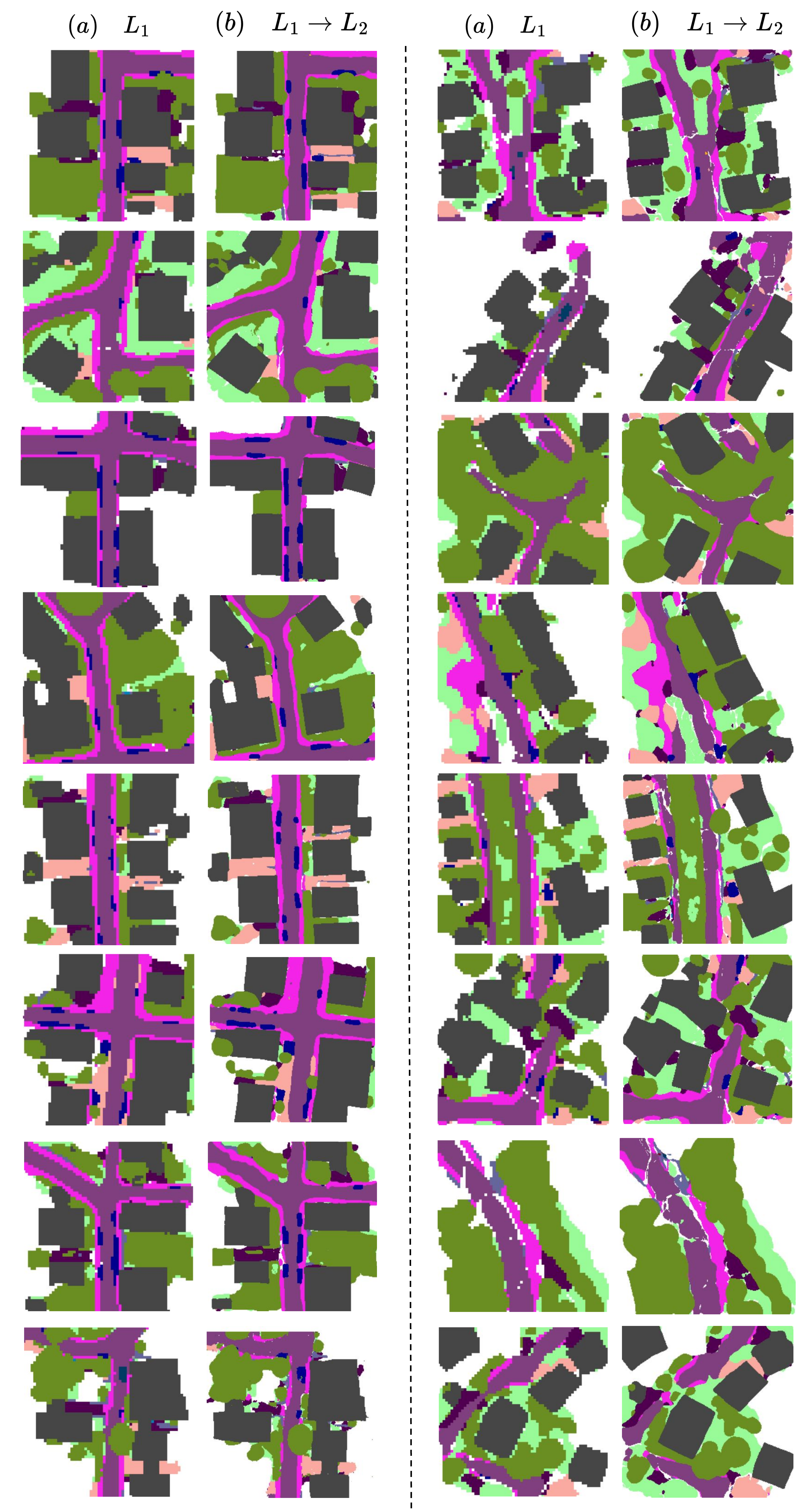}
\caption{\textbf{Hierarchical BEV Renderings of XCube-generated Scenes.} Each column pair shows results across successive hierarchy levels. Examples on the left demonstrate high-quality generations, while those on the right exhibit geometric distortions and semantic inconsistencies.}
\label{fig:xcube_bev}
\end{figure*}

\boldparagraph{3D Semantic Scene Generation} \figref{fig:generation} presents 3D semantic scenes generated by our method. By conditioning the LDM on discrete scene labels controlling vegetation density, we can produce scenes with (a) high, (b) medium, and (c) low vegetation. Additional BEV semantic renderings of synthesized scenes, used for computing the generative metrics reported in the main paper, are shown in \figref{fig:pritti_bev}. Notably, conditioning on vegetation density serves only as a minimal illustration of controllability through our class-conditional DiT. This conditioning scheme is general and can be naturally extended to other primitive categories (e.g., vehicles) or their combinations. To demonstrate this, we additionally trained models conditioned on vehicle density and joint vegetation–vehicle densities. Qualitative results are shown in \figref{fig:generation_vehicles} and \figref{fig:generation_vegetation_vehicles}, respectively. For comparison, we provide 3D and BEV rendering results for all baselines, generated unconditionally: SemCity (\figref{fig:semcity}, \figref{fig:semcity_bev}), PDD (\figref{fig:pdd_hierarchical}, \figref{fig:pdd_bev}), and XCube (\figref{fig:xcube_hierarchical}, \figref{fig:xcube_bev}). As seen in their BEV renderings and \figref{fig:baselines_failures}, these methods often produce scenes with limited semantic consistency and structural coherence, exhibiting blended geometries, irregular object shapes, and noisy semantic predictions. These artifacts reduce the perceived realism of the results, aligning with the lower quantitative generative performance in the main paper.

\boldparagraph{Scene Inpainting and Outpainting} \figref{fig:inpainting} and \figref{fig:outpainting} present additional scene inpainting and outpainting results produced by our method, respectively. As illustrated in \figref{fig:inpainting}, by appropriately setting the binary latent mask $\mathbf{m}$ (see~\eqnref{eq:z_composition}), we can achieve flexible, localized scene edits in arbitrary directions (\eg top, bottom, left, right). Among the baselines, only SemCity reports both inpainting and outpainting results. \figref{fig:applications} compares our method with SemCity, demonstrating that PrITTI produces coherent and semantically consistent edits and extensions, whereas SemCity often yields fragmented or contextually inconsistent structures, particularly in the outpainting setting. XCube does not address these scene-level editing tasks, and PDD provides only qualitative results for large-scale scene extrapolation, which we compare against in a later section. 

\boldparagraph{Ground-conditioned 3D Primitives Generation} \figref{fig:ground_to_objects} illustrates a special case of inpainting enabled by our disentangled latent representation. By keeping the ground channels fixed and inpainting only the object channels, our model generates novel object primitives conditioned on a consistent ground layout. This is achieved using a pre-trained LDM without any additional training. 

\begin{figure*}[t]
\centering
\includegraphics[width=\textwidth,height=0.95\textheight, keepaspectratio]{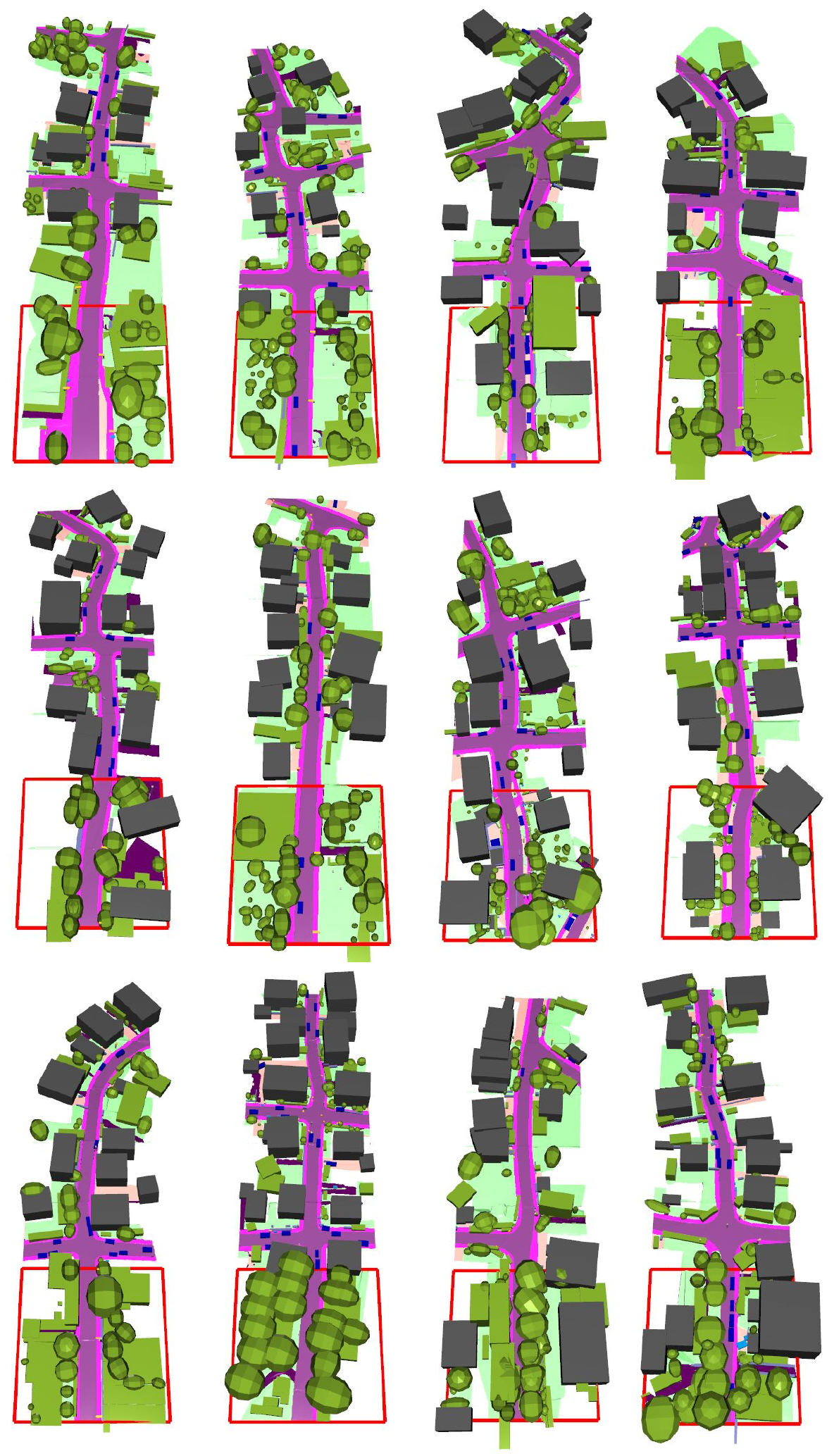}
\caption{\textbf{PrITTI Large-scale Scene Extrapolation Results.} Starting from an initially generated block (outlined in red), our method produces extended layouts that remain globally coherent and semantically consistent, with smooth structural transitions across blocks.}
\label{fig:progressive_outpainting}
\end{figure*}

\begin{figure*}[t]
\centering
\includegraphics[width=\textwidth,height=0.95\textheight, keepaspectratio]{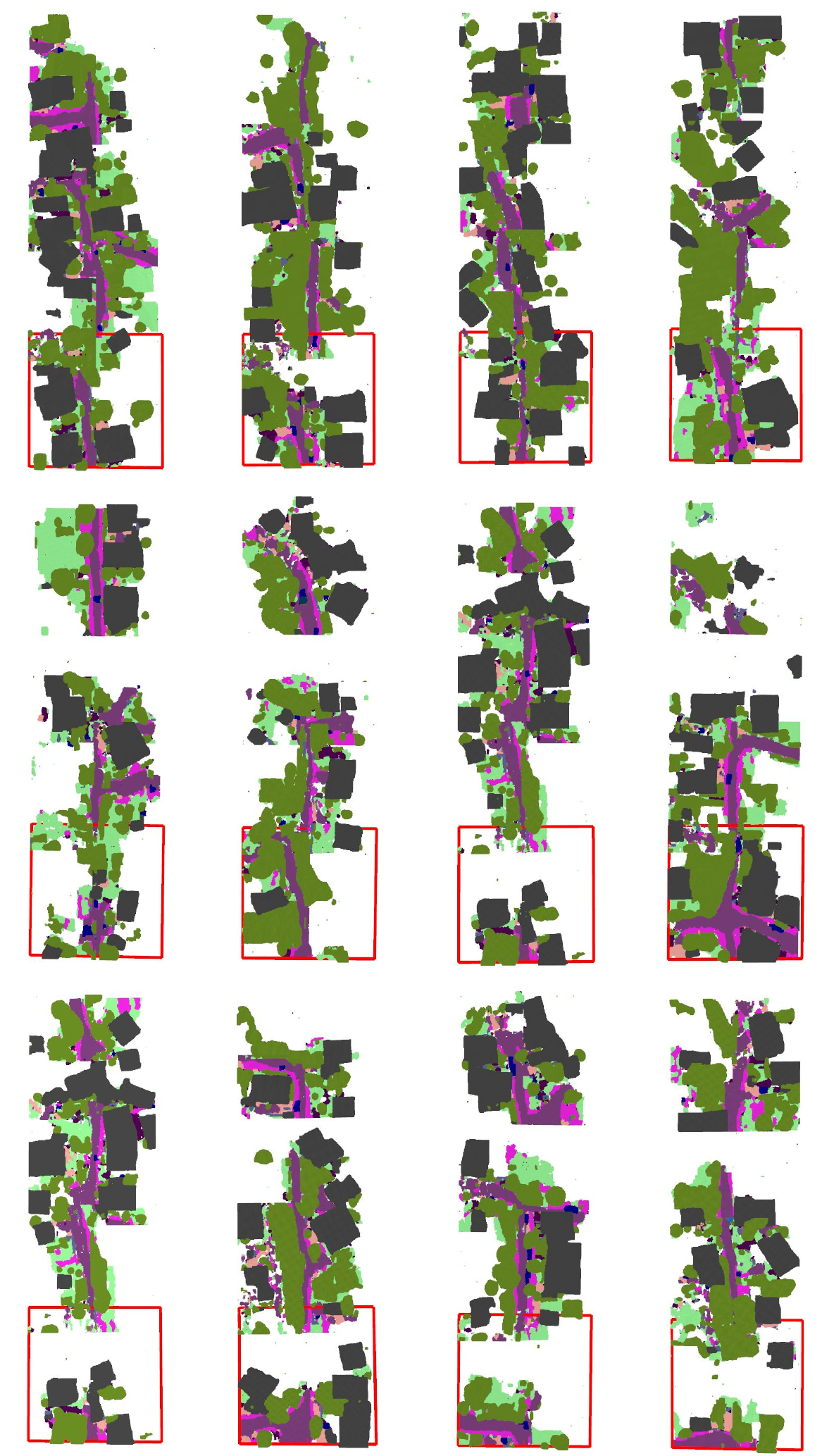}
\caption{\textbf{PDD Large-scale Scene Extrapolation Results.} Starting from an initially generated block (outlined in red), PDD progressively extends the scene upward. The extrapolated regions exhibit broken road connectivity, misplaced structures, and inconsistent semantic transitions, leading to fragmented large-scale layouts.}
\label{fig:pdd_progressive}
\end{figure*}

\boldparagraph{Large-scale Scene Extrapolation}
\figref{fig:progressive_outpainting} shows large-scale scene extrapolation results of our method, starting from an initially generated layout block (outlined in red) and progressively extending it upward. 
The extrapolated layouts exhibit strong global coherence, preserving smooth road connectivity and semantically consistent transitions between synthesized regions. All examples use the medium vegetation density condition. Changing the label to low or high can enable extrapolation of the same initial block under different vegetation settings, highlighting our model's controllability (see supplementary video for an example). For comparison, we also include results from PDD, the only baseline supporting large-scale scene extrapolation. SemCity provides code only for outpainting a generated block twice (see \figref{fig:applications}), and therefore cannot be evaluated here. As shown in~\figref{fig:pdd_progressive}, PDD often fails to maintain global structure: roads are disconnected between adjacent blocks, buildings are frequently misplaced, and abrupt semantic transitions occur across generated areas, resulting in overall fragmented and visually implausible large-scale layouts. In contrast, our method produces continuous, well-aligned extrapolations with consistent semantics and geometry across large regions.

\boldparagraph{Object-level Editing} Our framework maintains an instance-level vector representation for each primitive, parameterized by its 3D center location and 6D Cholesky parameters. This formulation allows intuitive manipulation of individual objects by directly updating their corresponding parameters (\eg center for translation, Cholesky parameters for rotation and scaling). \figref{fig:object_editing} illustrates additional editing results, where simple transformations are applied to randomly selected vehicle primitives from the original scene (column a). Columns (b-e) visualize examples of dropout, rotation, scaling, and translation, respectively. In contrast, object-level manipulation in voxel-based scene representations remains highly challenging. First, identifying which voxels correspond to a particular object typically requires explicit segmentation, which is often complicated by ambiguous object-background boundaries. Furthermore, even after segmentation, applying a transformation such as translation requires moving the corresponding voxel subset and then filling in the vacated space. Consequently, existing voxel-based approaches for 3D semantic urban scene generation provide little or no support for object-level editing. For instance, PDD and SemCity do not enable any form of direct instance-level manipulation, while XCube only supports manual voxel addition or removal for a single object at the coarsest level of its hierarchy through a dedicated user interface. 

\begin{figure*}[t]
\centering
\includegraphics[width=\textwidth,height=0.95\textheight, keepaspectratio]{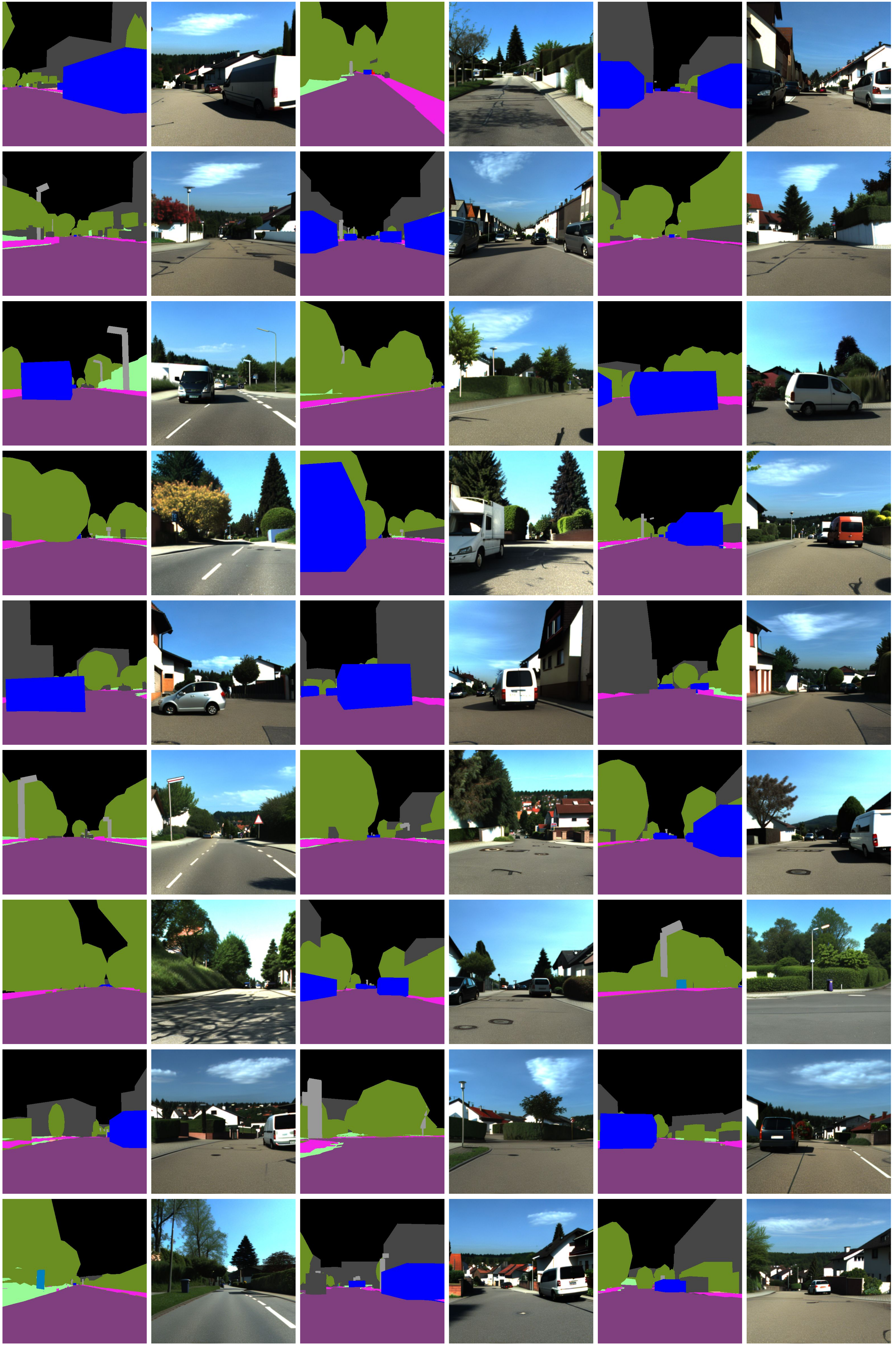}
\caption{\textbf{PrITTI Photo-realistic Street View Synthesis Results.} Each pair shows a 2D semantic rendering (left) and the corresponding photo-realistic image (right), generated with ControlNet~\cite{Zhang2023ICCVb}. Despite the coarse geometry of our primitives, the synthesized scenes exhibit diverse object appearances that go beyond cuboids or ellipsoids while remaining consistent with the input semantics.}
\label{fig:photorealistic}
\end{figure*}

\boldparagraph{Photo-realistic Street View Synthesis} \figref{fig:photorealistic} presents additional examples demonstrating how 2D semantic maps rendered from our generated 3D scenes can be translated into photo-realistic street-view images. For this task, we employ the ControlNet~\cite{Zhang2023ICCVb} model from Urban Architect~\cite{Lu2024ARXIV}, which is fine-tuned on KITTI-360 3D semantic layouts.  Although some visual artifacts exist, the rendered maps serve as effective conditioning signals and reliably guide the image generation process. Notably, despite the coarse geometry of our primitives, the synthesized images reflect object shapes beyond cuboids or ellipsoids. We emphasize that photo-realistic appearance synthesis is not the primary objective of our work but rather an additional downstream application.

\begin{figure*}[t]
  \centering
  \begin{subfigure}[t]{0.48\textwidth}
    \includegraphics[width=\textwidth]{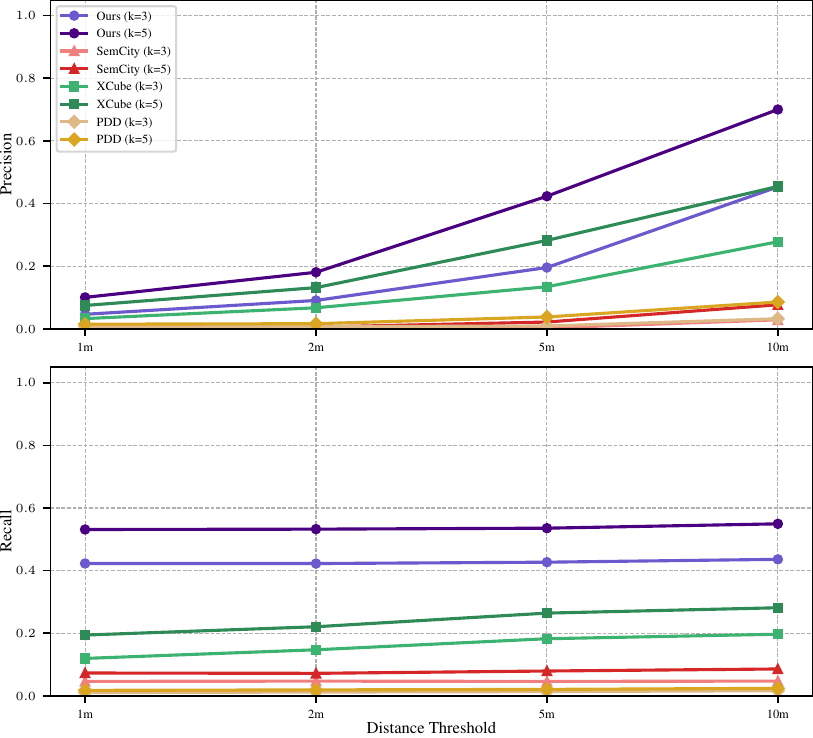}
    \caption{Distance Threshold Sampling}
    \label{fig:threshold_sampling}
  \end{subfigure}
  \hfill
  \begin{subfigure}[t]{0.48\textwidth}
    \includegraphics[width=\textwidth]{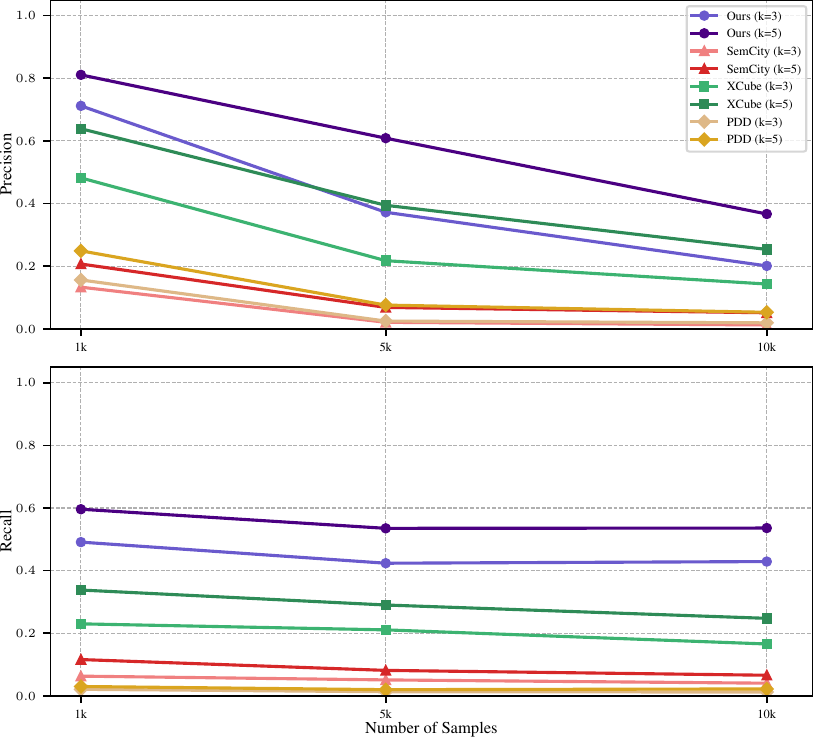}
    \caption{Farthest Pose Sampling}
    \label{fig:fps_sampling}
  \end{subfigure}
  \caption{\textbf{Precision and recall comparison} between our method (using DiT-B), SemCity, XCube, and PDD under two evaluation protocols: (a) distance-threshold reference-pose sampling and (b) farthest-pose sampling, each tested across varying neighborhood sizes.}
  \label{fig:sampling_strategy}
\end{figure*}

\section{Precision and Recall Metrics}
To evaluate our generative models, among others, we use the improved precision and recall metrics introduced by Kynkäänniemi et al.~\cite{Kynkaanniemi2019NeurIPS}. Let \( \boldsymbol{\Phi}_r \) and \( \boldsymbol{\Phi}_g \) denote the feature sets extracted from real and generated samples, respectively. A binary function \( f(\boldsymbol{\phi}, \boldsymbol{\Phi}) \) determines whether a feature vector \( \boldsymbol{\phi} \) lies within the estimated manifold of a reference set \( \boldsymbol{\Phi} \in \{ \boldsymbol{\Phi}_r, \boldsymbol{\Phi}_g \} \):

\begin{equation}
f(\boldsymbol{\phi}, \boldsymbol{\Phi}) =
\begin{cases}
1, &
\text{if } 
\|\boldsymbol{\phi} - \boldsymbol{\phi}'\|_2
\leq
\|\boldsymbol{\phi}' - \text{NN}_k(\boldsymbol{\phi}', \boldsymbol{\Phi})\|_2, \\[2pt]
&\text{for at least one } \boldsymbol{\phi}' \in \boldsymbol{\Phi}, \\[4pt]
0, & \text{otherwise.}
\end{cases}
\label{eq:binary_function}
\end{equation}

\noindent Here, \( \text{NN}_k(\boldsymbol{\phi}', \boldsymbol{\Phi}) \) denotes the \(k\)\textsuperscript{th} nearest neighbor 
of \( \boldsymbol{\phi}' \) within \( \boldsymbol{\Phi}\). Using this function, precision and recall are computed as:

\begin{align}
\text{precision}(\boldsymbol{\Phi}_r, \boldsymbol{\Phi}_g) = \frac{1}{|\boldsymbol{\Phi}_g|} \sum_{\boldsymbol{\phi}_g \in \boldsymbol{\Phi}_g} f(\boldsymbol{\phi}_g, \boldsymbol{\Phi}_r) \\
\text{recall}(\boldsymbol{\Phi}_r, \boldsymbol{\Phi}_g) = \frac{1}{|\boldsymbol{\Phi}_r|} \sum_{\boldsymbol{\phi}_r \in \boldsymbol{\Phi}_r} f(\boldsymbol{\phi}_r, \boldsymbol{\Phi}_g)
\label{eq:precision_recall}
\end{align}

\noindent Intuitively, precision measures the proportion of generated samples that appear realistic (\ie fall within the manifold of real data), whereas recall quantifies how well the generated samples cover the diversity of real samples. Following~\cite{Kynkaanniemi2019NeurIPS}, we use an equal number of real and generated samples (\( |\boldsymbol{\Phi}_r| = |\boldsymbol{\Phi}_g| \)). However, we found out that these metrics are highly sensitive to several factors including: \textbf{(i)} implementation details, \textbf{(ii)} the diversity of real samples, \textbf{(iii)} the number of evaluation samples, and \textbf{(iv)} the neighborhood size \( k \) used for manifold estimation (see~\eqnref{eq:binary_function}). For \( \textbf{(i)}\), we follow~\cite{Peebles2023ICCV}, which forms the basis of our DiT models, and use the evaluation suite from~\cite{Dhariwal2021NeurIPS} for consistency. To address \( \textbf{(ii)}\), we implement two separate strategies for sampling real poses. The first enforces a minimum spatial distance (in meters) between neighboring poses. The second, inspired by farthest point sampling (FPS), selects poses sequentially to maximize their distance from previously chosen ones. The number of evaluation samples \textbf{(iii)} depends on the applied sampling strategy.
In the distance-based approach, the sample size is implicitly determined by the distance constraint, whereas in FPS we explicitly control it, evaluating with 1K, 5K, and 10K real samples. Finally, to study \textbf{(iv)}, we vary the neighborhood size \(k \in \{3,5\}\), examining its effect on the generative performance. Comprehensive evaluations of precision and recall under these configurations are presented in \figref{fig:sampling_strategy}, comparing our approach against SemCity, XCube, and PDD. For PDD and XCube, results are reported at their finest resolution, and for SemCity we use the 1M variant. %

For both sampling strategies, we observe a consistent decrease in precision as the number of evaluation samples increases. In the distance-based strategy, this effect corresponds to smaller distance thresholds (\eg \( 1\,m\) yields 32,203 samples, whereas \( 10\,m\) yields 3,855). This behavior follows~\eqnref{eq:binary_function}: as more real samples are included, the hyperspheres defined by their \(k\)\textsuperscript{th} nearest neighbors become smaller and more densely packed. Consequently, even visually realistic generated samples are less likely to fall within any real sample's hypersphere. In contrast, recall remains relatively stable across varying sample counts, likely because denser sampling refines the real manifold without altering its overall extent. Finally, both precision and recall increase with larger neighborhood sizes \(k\). This trend follows directly from~\eqnref{eq:binary_function}: increasing \(k\) expands each sample's hypersphere, thereby loosening the inclusion criterion.

\section{Overfitting Analysis}
\begin{figure}[t]
  \centering
  \includegraphics[width=\linewidth]{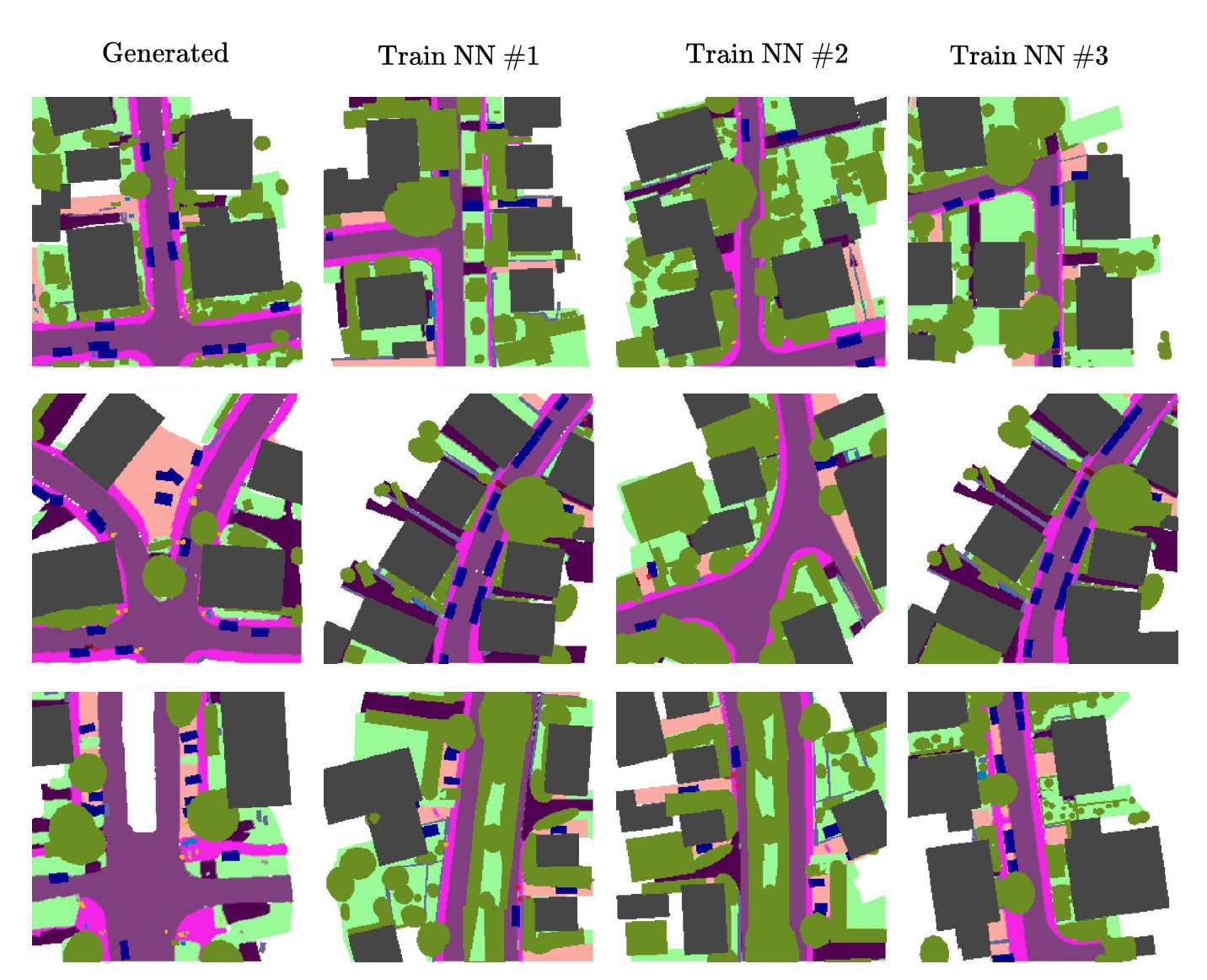}
  \caption{\textbf{Nearest-neighbor Analysis.} For each generated sample (left), we show its closest training samples (right). Generated scenes differ in layout and object placement, suggesting the model does not memorize training data.}
  \vspace{-0.5cm}
  \label{fig:overfitting_analysis}
\end{figure}

To assess potential memorization of dataset-specific patterns, we perform a nearest-neighbor (NN) analysis in feature space. We extract features using a pretrained VGG16 network and compute Euclidean NN distances between 50K generated BEV samples and a fixed subset of 50K training samples. To contextualize these values, we additionally compute real-to-real distances on the same subset by matching each training sample to its nearest neighbor excluding self-matches. We observe that generated samples are consistently farther from the training set (mean NN distance $\approx 52.2$, median $\approx 51.7$, minimum $\approx 27.2$, max $\approx 115.7$) than real samples are from each other (mean NN distance $\approx 22.5$, median $\approx 22.0$, minimum $\approx 4.9$, max $\approx$ 58.7). This suggests that the model does not simply memorize or reproduce training examples. This is further supported by the qualitative examples in~\figref{fig:overfitting_analysis}, where generated samples differ from their three nearest neighbors in the training set in layout and object arrangement.

\begin{figure*}[t]
\centering
\includegraphics[width=\textwidth]{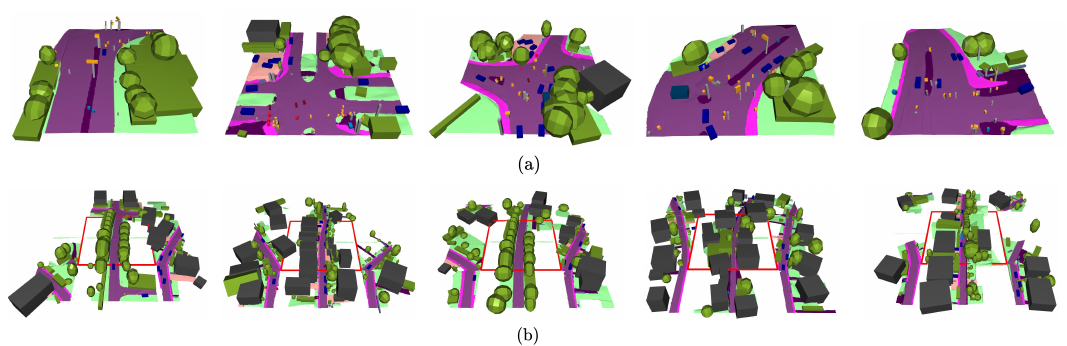} \\
\caption{\textbf{PrITTI Failure Cases.} (a) Fine-grained attachments (\eg pole-mounted lamps) are often not accurately reconstructed, yielding floating objects. (b) Large empty conditioning regions in outpainting can lead to weak or inconsistent completions.}
\label{fig:limitations}
\end{figure*}

\section{Applicability beyond KITTI-360} We conduct our main experiments on KITTI-360, which, to the best of our knowledge, is the only real-world dataset providing primitive-level annotations for entire urban scenes (\eg buildings, vegetation) rather than focusing on selected object classes (\eg vehicles). This yields scenes with many primitives and thus a demanding setting for training and evaluation. Importantly, PrITTI does not require \textit{dense} primitive annotations, but only parametric 3D cuboids or ellipsoids for arbitrary semantic categories. When fewer categories are annotated, the model simply predicts fewer primitives overall, which makes the task easier. Many autonomous driving datasets~\cite{Caesar2021CVPR,Caesar2021CVPRW,Argoverse22021NeurIPS} already provide such cuboid annotations. Training PrITTI on any of them involves grouping their labels into dataset-specific primitive categories and converting the cuboid parameters into our vectorized representation. For ground geometry, 3D HD maps (\eg in~\cite{Argoverse22021NeurIPS}) enable BEV height fields rendering, while 2D maps (\eg in~\cite{Caesar2021CVPR}) can be handled via a flat-ground assumption. 

\begin{figure*}[t]
\centering
\vspace{-0.5cm}
\includegraphics[width=0.8\textwidth]{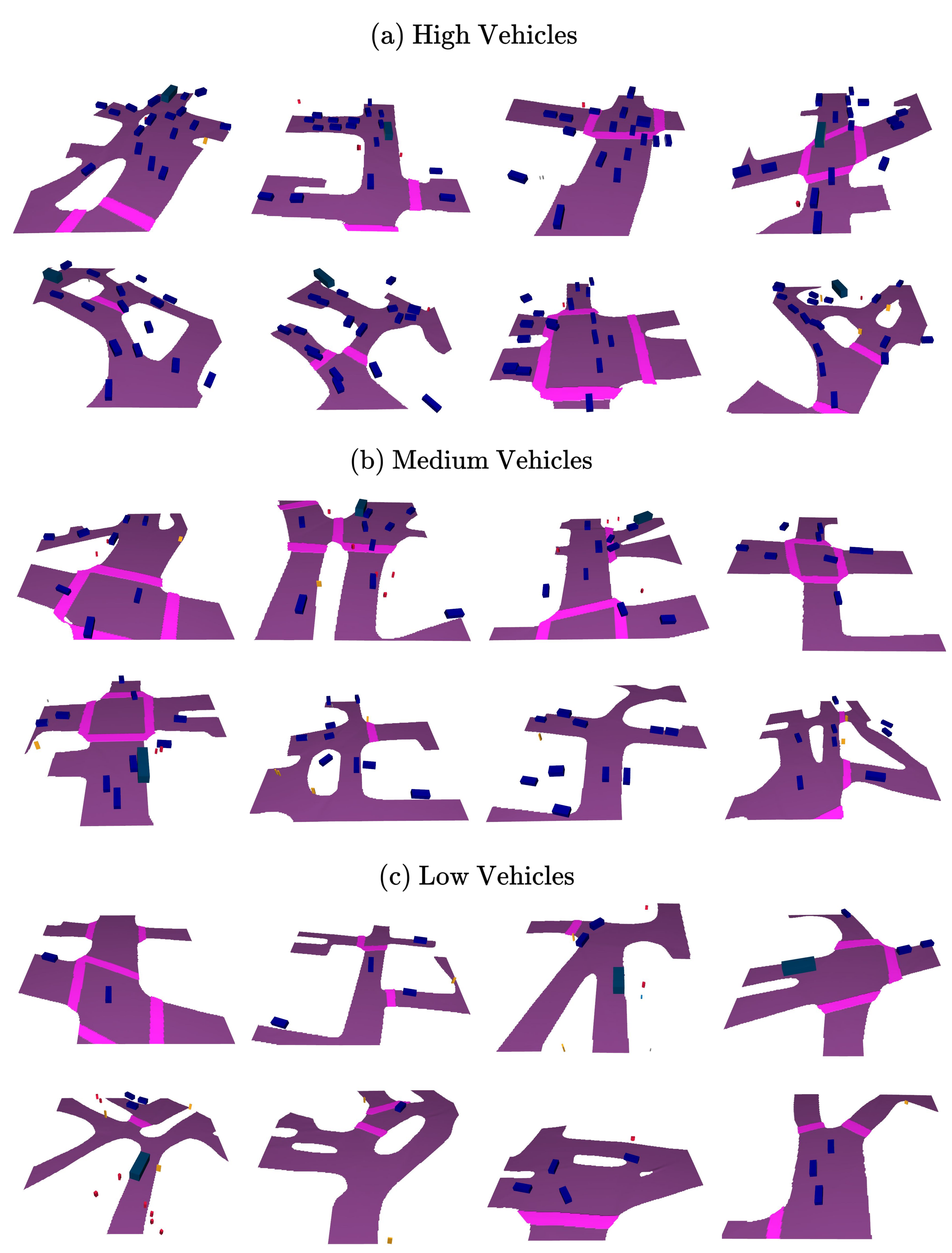} \\
\caption{\textbf{PrITTI Scene Generation Results on Argoverse 2 (AV2) Dataset.} Generated 3D semantic scenes grouped under three vehicle-density conditions: (a) high, (b) medium, and (c) low. Vehicles appearing outside the drivable surface correspond to parked cars, as AV2~\cite{Argoverse22021NeurIPS} HD maps do not provide explicit parking area annotations, unlike KITTI-360.}
\label{fig:av2_generation}
\end{figure*}

To demonstrate applicability beyond KITTI-360, we additionally train our method on Argoverse 2 (AV2)~\cite{Argoverse22021NeurIPS}. We group the original AV2 labels into seven primitive categories: vehicle big, vehicle small, two-wheelers, human, pole, traffic control, and object. We emphasize that PrITTI does not require a fixed category set across datasets, and for AV2 these groups happen to align well with a subset of those used in KITTI-360 (see~\tabref{tab:object_categories}). Notably, we find from dataset statistics that only 84 total queries are required, compared to 514 for~\cite{Liao2022PAMI}, highlighting the reduced task complexity. We report qualitative generation results in~\figref{fig:av2_generation} and quantitative results for both reconstruction and generation in~\tabref{tab:av2_results}.

\begin{table}[t]
\centering
\small
\setlength{\tabcolsep}{6pt}

\begin{subtable}{\linewidth}
\centering
\begin{tabularx}{\linewidth}{
    >{\centering\arraybackslash}X
    >{\centering\arraybackslash}X
    |
    >{\centering\arraybackslash}X
    >{\centering\arraybackslash}X
}
\toprule
\multicolumn{2}{c|}{Raster ($\times 10^{-2}$)} & \multicolumn{2}{c}{Primitive} \\
\cmidrule(r){1-2} \cmidrule(l){3-4}
\textbf{MSE} $\downarrow$ & \textbf{IoU} $\uparrow$ & \textbf{AP3D} $\uparrow$ & \textbf{AP3D@50} $\uparrow$ \\
\midrule
0.015 & 99.98 & 56.495 & 39.490 \\
\bottomrule
\end{tabularx}
\caption{Reconstruction}
\end{subtable}

\vspace{0.8em}

\begin{subtable}{\linewidth}
\centering
\begin{tabularx}{\linewidth}{
    >{\centering\arraybackslash}X
    >{\centering\arraybackslash}X
    >{\centering\arraybackslash}X
    >{\centering\arraybackslash}X
}
\toprule
\textbf{Precision} $\uparrow$ & \textbf{Recall} $\uparrow$ & \textbf{FID} $\downarrow$ & \textbf{IS} $\uparrow$ \\
\midrule
0.572 & 0.410 & 144.330 & 2.878 \\
\bottomrule
\end{tabularx}
\caption{Generation (DiT-B)}
\end{subtable}

\caption{PrITTI evaluation results on AV2~\cite{Argoverse22021NeurIPS}, showing (a) reconstruction and (b) generation metrics.}
\label{tab:av2_results}
\end{table}

\section{Limitations and Failure Cases}
\begin{figure*}[t]
\centering
\includegraphics[width=\textwidth,height=0.95\textheight, keepaspectratio]{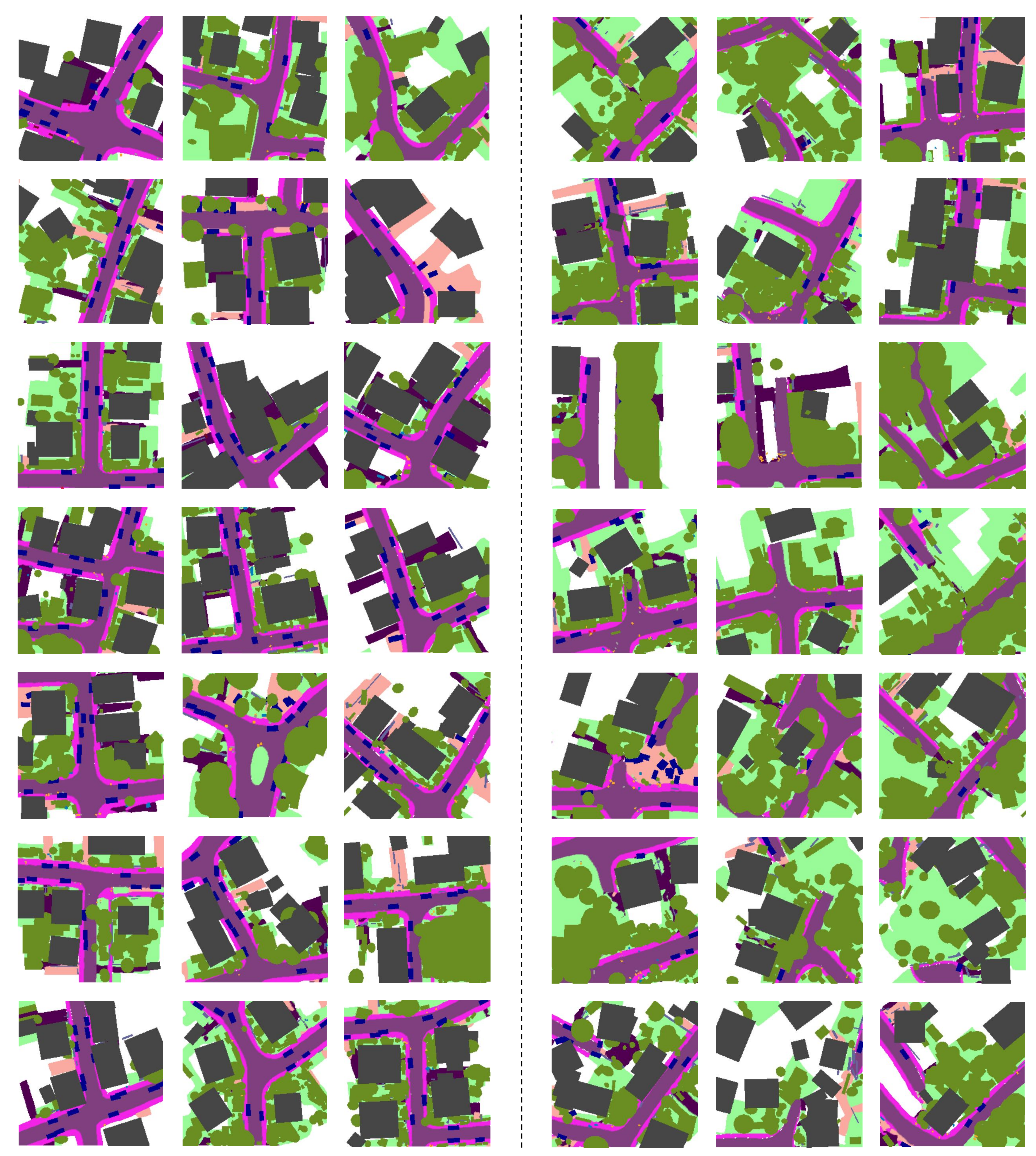}
\caption{\textbf{BEV Renderings of PrITTI-generated Scenes.} Examples on the left illustrate well-structured and realistic scene generations, while those on the right show poorly synthesized samples with geometric or semantic inconsistencies.}
\label{fig:pritti_bev}
\end{figure*}

Despite its promising results, our approach comes with certain limitations. First, as discussed in the main paper, the current framework is limited to static scenes and represents objects using coarse 3D primitives, which reduces geometric fidelity. Second, the model operates on a fixed set of semantic categories. Although this aligns with prior semantic layout methods~\cite{Lee2024CVPR,Zheng2024ARXIV,Liu2024ECCV,Ren2024CVPR}, it limits the generation of unseen object types. Extending the framework toward open-vocabulary scene generation is an exciting future direction. In addition, ground elements are modeled as rasterized 2D representations that are subsequently extruded into 3D. Compared to voxel-based approaches that maintain a 3D grid over all scene elements, using a 2D representation yields better memory and computational scalability. However, the extrusion process is still inherently constrained by the resolution of the input raster maps, potentially resulting in less detailed ground surfaces. Furthermore, accurately reconstructing fine-grained attachments (\eg lamps or traffic lights mounted on poles) and small, densely arranged structures (\eg fences) remains quite challenging. Such geometric inaccuracies can be learned by the autoencoder and propagated to the diffusion model, occasionally producing scenes with floating or disconnected objects (see \figref{fig:limitations} (a)). Moreover, our framework does not explicitly enforce hard geometric constraints %
and instead relies on training data to capture such relationships. While the generated scenes are generally plausible, occasional geometric inconsistencies arise, especially in crowded scenes. In particular, primitives may intersect, which can sometimes be valid (\eg vehicles with attached trailers) but not always appropriate (\eg two vehicles placed unrealistically close to each other). Similarly, ground-primitive misalignment can produce semantically inconsistent configurations, such as buildings intersecting the drivable road surface (~\figref{fig:pritti_bev}, right). Incorporating soft constraints or geometry-aware guidance during diffusion~\cite{Lu2024ICRA} could mitigate these issues and improve physical validity. Finally, in the outpainting setting, failures may occur when the conditioning block %
contains large empty regions along certain extrapolation directions. In such cases, the diffusion model is afforded greater generative freedom, which can lead to weakly connected or semantically inconsistent completions (see \figref{fig:limitations} (b)). Additionally, some visible seams can appear across block transitions in the ground surfaces.

\section{Potential Societal Impacts}
Our work addresses 3D semantic scene generation for urban environments using primitive-based representations. Traditional approaches to constructing such 3D environments require expert knowledge and significant manual effort, making them costly and time-consuming. Moreover, reliance on pre-existing layout data, which remains scarce, particularly for outdoor scenes, limits diversity and leads to repetitive scene compositions. In contrast, our method enables the automatic generation of diverse and realistic 3D semantic scenes. This has the potential to benefit a range of applications in simulation, robotics, and autonomous driving. By providing interpretable and controllable 3D content, it can contribute to safer and more scalable environments for testing autonomous systems and improved scene understanding for embodied agents. Nevertheless, as with all generative models, there is potential for misuse. Synthetic scenes could, for instance, be employed to misrepresent real urban layouts, simulate favorable outcomes in planning proposals, or test vision-based monitoring systems (\eg for traffic detection) under unrealistic conditions. Furthermore, our primitive-based representation simplifies real-world geometry, which, if over-relied upon, may lead to misleading conclusions in downstream tasks. These risks, however, do not diminish the framework's practical value. As with existing modeling and simulation tools, effective use relies on a clear understanding of the method's assumptions and limitations. 

\begin{figure*}[t]
\centering
\includegraphics[width=\textwidth,height=0.95\textheight, keepaspectratio]{supp/gfx/21_vehicle_conditioned_generation.pdf}
\caption{\textbf{PrITTI Vehicle Density-Conditioned Scene Generation.} Generated 3D semantic scenes grouped under three vehicle-density conditions: (a) high, (b) medium, and (c) low, demonstrating controllability over primitive categories beyond vegetation.}
\label{fig:generation_vehicles}
\end{figure*}

\begin{figure*}[t]
\centering
\includegraphics[width=\textwidth,height=0.95\textheight, keepaspectratio]{supp/gfx/22_vegetation_vehicle_conditioned_generation.pdf}
\caption{\textbf{PrITTI Joint Vegetation–Vehicle Density-Conditioned Scene Generation.} Generated scenes conditioned on combinations of vegetation and vehicle densities. From the nine possible settings, we show three examples: (a) high vegetation, low vehicles, (b) medium vegetation, medium vehicles, and (c) low vegetation, high vehicles, illustrating controllability over combinations of primitive categories.}
\label{fig:generation_vegetation_vehicles}
\end{figure*}

\begin{figure*}[t]
\centering
\includegraphics[width=\textwidth,height=0.95\textheight, keepaspectratio]{supp/gfx/08_inpainting.pdf}
\caption{\textbf{PrITTI Inpainting Results.} Inpainting applied to (a) top, (b) bottom, (c) right, and (d) left regions of an input scene. PrITTI produces localized edits that blend seamlessly with the surrounding geometry and semantics, yielding consistent and plausible completions.}
\label{fig:inpainting}
\end{figure*}

\begin{figure*}[t]
\centering
\includegraphics[width=\textwidth,height=0.95\textheight, keepaspectratio]{supp/gfx/19_ground_to_objects.pdf}
\caption{\textbf{PrITTI Ground-to-Object Inpainting Results.} Leveraging our disentangled joint latent structure, we can inpaint object primitives while keeping the ground layout fixed. The results show that the ground geometry is faithfully preserved from the original scene, while newly generated objects are placed in semantically appropriate locations.}
\label{fig:ground_to_objects}
\end{figure*}

\begin{figure*}[t]
\centering
\includegraphics[width=\textwidth,height=0.95\textheight, keepaspectratio]{supp/gfx/02_applications.pdf}
\caption{\textbf{Comparison of scene inpainting and outpainting results} on identical test scenes using our method and SemCity~\cite{Lee2024CVPR}. PrITTI (Ours) yields coherent and semantically consistent edits and extensions, while SemCity often produces fragmented or contextually inconsistent structures, particularly during outpainting.}
\label{fig:applications}
\end{figure*}

\begin{figure*}[t]
\centering
\includegraphics[width=\textwidth,height=0.95\textheight, keepaspectratio]{supp/gfx/14_outpainting.pdf}
\caption{\textbf{PrITTI Outpainting Results.} Outpainting examples produced by our method showing that it can expand a scene beyond its spatial extent while preserving ground layout continuity and generating new objects that integrate naturally with the surrounding context.}
\label{fig:outpainting}
\end{figure*}

\begin{figure*}[t]
\centering
\includegraphics[width=\textwidth,height=0.95\textheight, keepaspectratio]{supp/gfx/09_object_editing.pdf}
\caption{\textbf{PrITTI Object Editing Results.} Enabled by our instance-level primitive representation, PrITTI naturally supports a wide range of individual-object manipulations. Columns (b-e) show dropout, rotation, scaling, and translation applied to the same randomly selected vehicle primitives from (a), highlighted in white.}
\label{fig:object_editing}
\end{figure*}

\end{document}